\documentclass[10pt,twocolumn,letterpaper]{article}

\usepackage{iccv}
\usepackage{times}
\usepackage{epsfig}
\usepackage{graphicx}
\usepackage{amsmath}
\usepackage{amssymb}
\usepackage{subfigure}

\usepackage[pagebackref=true,breaklinks=true,letterpaper=true,colorlinks,bookmarks=false]{hyperref}

\newcommand{\argmax}{\mathop{\rm arg~max}\limits}
\iccvfinalcopy 

\def\iccvPaperID{327} 

\ificcvfinal\fi
\begin{document}

\title{TransCut: Transparent Object Segmentation from a Light-Field Image}

\author{Yichao Xu, Hajime Nagahara, Atsushi Shimada, Rin-ichiro Taniguchi\\
Kyushu University, Japan\\
{\tt\small \{xu,nagahara,atsushi,rin\}@limu.ait.kyushu-u.ac.jp}
}

\maketitle

\begin{abstract}
The segmentation of transparent objects can be very useful in computer vision applications. However, because they borrow texture from their background and have a similar appearance to their surroundings, transparent objects are not handled well by regular image segmentation methods. We propose a method that overcomes these problems using the consistency and distortion properties of a light-field image. Graph-cut optimization is applied for the pixel labeling problem. The light-field linearity is used to estimate the likelihood of a pixel belonging to the transparent object or Lambertian background, and the occlusion detector is used to find the occlusion boundary.  We acquire a light field dataset for the transparent object, and use this dataset to evaluate our method. The results demonstrate that the proposed method successfully segments transparent objects from the background.
\end{abstract}

\section{Introduction}
Image segmentation is a fundamental problem in computer vision. The goal of segmentation is to simplify and/or change the representation of an image into something that is more meaningful and easier to analyze \cite{shapiro2001computer}.
For example, it is very important to separate foreground objects from the background in applications such as object detection, object recognition \cite{mori2004recovering}, and surveillance tasks \cite{conaire2006multispectral}.
Numerous methods have been developed to deal with the image segmentation problem, including techniques based on thresholding \cite{otsu1975threshold}, partial differential equations \cite{caselles1997geodesic}, and graph partitioning \cite{shi2000normalized, boykov2006graph}.
However, none of these methods are suitable for the segmentation of transparent objects from an image. The difficulty of dealing with such objects means that transparent object segmentation is a relatively untouched field.

Although there are few techniques for separating transparent objects from an image, many tasks in our everyday life deal with transparent objects. For example, when a machine is operating in kitchens, living rooms, and offices, it should avoid touching fragile objects such as glasses, vases, bowls, bottles, and jars. One way of detecting these transparent objects is to segment them from captured images of the scene. The appearance of a transparent object is highly dependent on the background, from which its texture and colors are largely borrowed. Thus, it is extremely challenging to separate the transparent object from the background.

\begin{figure}[t]
	\centering
	\includegraphics[width=.8\columnwidth]{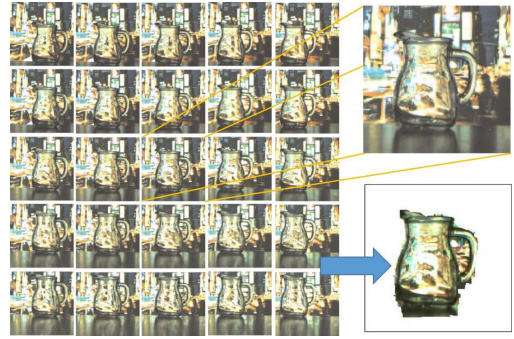}
	\caption{Input and output of our system. The left side shows the captured light-field image, and the right-hand side is a magnification of the central viewpoint. The output after segmentation of the transparent object is shown on the bottom-right.}
    \label{fig:input_output}
\end{figure}
\begin{figure}[t]
    \centering
	\subfigure[Linear features from Lambertian area.]{
		\includegraphics[clip, width=0.8\columnwidth]{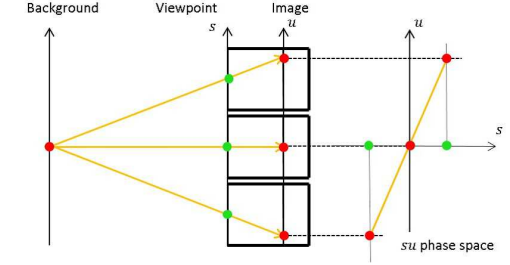}
		\label{fig:optical_path}
	}
	\subfigure[Non-linear feature from transparent object.]{
		\includegraphics[clip, width=0.8\columnwidth]{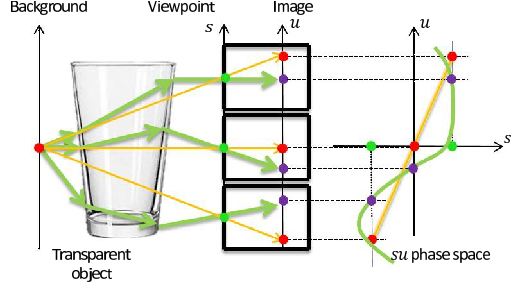}
		\label{fig:optical_path_distorted}
	}
	\caption{Different properties of the LFD feature.}
	\label{fig:LF_propagation}
\end{figure}

It is almost impossible to achieve stable transparent object segmentation in a 2D image using conventional image segmentation approaches.
In this paper, we utilize a light-field camera to capture 4D light-field images, and propose a method that can segment the transparent objects from the captured 4D light-field image (see Fig. \ref{fig:input_output}). Our method can automatically segment the transparent objects without any interaction.
The main idea is to take advantage of the light-field distortion (LFD) feature \cite{trans_recog} that has been proposed for transparent object recognition. LFD does not rely on the appearance of the background, and LFD features from the Lambertian and non-Lambertian areas have different properties. As shown in Fig. \ref{fig:LF_propagation}, the features from the Lambertian area are almost linearly distributed with respect to the viewpoints, unlike features from transparent objects. We call this property light-field linearity (LF-linearity).
However, features from the background will be nonlinear when occlusion occurs, and features from the transparent object will be linear when the distortion is relatively mild.
This does not present a problem when LFD is used for recognition tasks, because several dominant features can determine what type of transparent object is contained in the image. We cannot completely separate the transparent object from the background using LFD alone, but we can use this feature to obtain a rough estimate of the position of the transparent object and the background.

To completely segment the transparent object from an image, we utilize the graph-cut optimization method \cite{boykov2006graph} with LF-linearity and occlusion detection from the 4D light-field image. Our method only uses information from a geometric relationship that is independent of the color and texture.

The contributions of this paper are as follows: 1) we propose a method for a challenging computer vision problem, transparent object segmentation, and the method is automatic, requiring no human interaction; 2) an energy function is defined using the LF-linearity, and occlusion detector; and 3) comparisons show that the proposed method obtains better results than previous method for finding glass \cite{findingGlass}.


\section{Related work}\label{sec:relatedWork}
There are various strategies for optimizing energy functions. The combinatorial min-cut/max-flow graph-cut algorithm is widely used for energy functions defined on a discrete set of variables.
Greig et al. \cite{greig1989exact} were the first to realize that powerful min-cut/max-flow algorithms could be used to minimize certain energy functions in computer vision applications.
The energy function encodes both regional object information and the regularization of the image smoothness.
The regional information usually comes from user interaction \cite{boykov2006graph, rother2004grabcut}, particularly in image editing applications.
Automatic segmentation approaches that do not require user interaction have been developed in recent years.
An object segmentation framework \cite{carreira2010constrained} has been proposed for the automatic extraction of candidate objects by solving a sequence of constrained parametric min-cut problems.
Another method \cite{papazoglou2013fast} estimates whether a pixel is inside the foreground object based on the point-in-polygon problem, whereby any ray starting from a point inside the polygon will intersect the boundary of the polygon an odd number of times.
In our method, we use occlusion to detect the boundary of a transparent object, and this occlusion boundary also allows us to determine which side is the background. We detect the occlusion boundary by designing a series of occlusion detectors to check the pattern of forward-backward matching consistency in all viewpoints. The forward-backward matching consistency has been used in many previous studies such as \cite{alvarez2007symmetrical}. For more sophisticated occlusion detection strategies, we refer to \cite{ayvaci2012sparse} and the references therein.

From a different perspective, a number of studies have analyzed images captured by special optics or devices to obtain the physical parameters of the transparent object, such as its refractive index and surface normal.
Schlieren photography \cite{Settles01,Settles10} has been used to analyze gas and fluid flows and shock waves.
This method requires high-quality and precisely aligned optics to visualize the refraction response in a scene as a gray-scale or color image.
Wetzstein et al. \cite{Wetzstein11} extended this technique to light-field background-oriented Schlieren photography, using a common digital camera and a special optical sheet, known as a light-field probe (LF-probe), to reconstruct the transparent surface \cite{Wetzstein11b}. Similarly, Ji et al. \cite{ji2013reconstructing} used an LF-probe and multiple viewpoints to reconstruct an invisible gas flow.
The light refracted by transparent objects tends to be polarized, meaning that polarizing filters can be used to measure their light intensity \cite{Miyazaki05, Miyazaki04}.
Ding et al. \cite{ding2011dynamic} used a camera array and checkerboard pattern to acquire dynamic 3D fluid surfaces. Ye et al. \cite{ye2012angular} acquired dynamic 3D fluid surfaces with a single camera, but they used a special "Bokode" background (which emulates a pinhole projector) to capture ray--ray correspondences.
All of the above methods require some special optics or devices, so their applicability is restricted to laboratory environments, and they are not feasible for common practical use.

Similar to our target, learning-based method \cite{findingGlass2, findingGlass} has been proposed for finding glass in a single view image.
Fritz et al. \cite{Fritz09} used SIFT feature and LDA for learning a transparent object and detecting its location and region as a bounding box.
Wang et al. \cite{wang2012glass,wang2013glass} used RGB-D image for glass object segmentation. The depth image was utilized as one of the cues for transparency that the depth information is missing in the glass region, since the glass refracts the active light from the sensor.
For multi-view images as input, the epipolar-plane-image (EPI) analysis method was used to extract layers with specular properties \cite{Criminisi2005}.
Multi-view images with known camera motion has been used to recover shape and pose of transparent object \cite{Ben-ezra03}.
Our method utilize the characteristic that transparent objects distort the background by refraction to derive the LFD feature \cite{trans_recog} which was originally proposed for transparent object recognition.
We take advantage of occlusion information and the distortion feature for transparent object segmentation from a single-shot light-field image, and the proposed method also has the potential for glass and specular objects.
\section{Feature descriptors from light field}\label{sec:LFC_LFD}
In this section, we define LF-linearity and occlusion detector for describing feature of transparent object.
\subsection{Light-field linearity} \label{sec:LF-linearity}
The LFD feature was proposed by Maeno et al. \cite{trans_recog}. They used this feature to classify different shapes of transparent objects. We will utilize an important property of this feature to the likelihood of a pixel being the Lambertian background.

Similar to \cite{trans_recog}, we define the LFD as the set of relative differences between the coordinates of the corresponding points:
\begin{align}
\mathbf{LFD}(u,v)=\{(s,t,\Delta u, \Delta v) |(s,t)\neq (0,0) \},
\label{eq:lfd}
\end{align}
where $(s,t)$ is the viewpoint coordinate, and $(\Delta u, \Delta v)$ is the difference between point $p(0,0,u,v)$ in the central viewpoint $view(0,0)$ and its corresponding point $p'(s,t,u',v')$ in viewpoint $view(s,t)$:
\begin{equation}\label{eq:delta_uv}
\left\{
\begin{aligned} 
\Delta u & = u'-u\\
\Delta v & = v'-v
\end{aligned}
\right.
\end{equation}
In the experiments, we use an optical flow algorithm to obtain the correspondences between the central viewpoint $view(0,0)$ and viewpoints $view(s,t)$.

As described in Fig. \ref{fig:LF_propagation}, the disparities in a transparent object include the refraction effect. Thus, the LFD features coming from the transparent object are more distorted than features from the background, and these features deviate from the hyperplane given by the Lambertian reflection in the phase space.
The hyperplane in the $stuv$-space containing point $p(0,0,u,v)$ can be described as:
\begin{equation}
n_1 s + n_2 t + n_3 \Delta u + n_4 \Delta v = 0,
\end{equation}
where $(s,t,\Delta u,\Delta v)$ is as before, i.e., the viewpoint coordinates and the difference between the corresponding image points. The positions of the viewpoints can be obtained by camera array calibration \cite{xu_qcav}.
$(n_1, n_2, n_3, n_4)$ is the unit normal vector $\vec{\mathbf{n}}$ of the hyperplane.
This vector is estimated by fitting $(s, t, \Delta u, \Delta v)$ from all $M$ viewpoints:
\begin{equation}
\underbrace{\left[
  \begin{array}{c}
	(~s, ~t, ~\Delta u, ~\Delta v)_1\\
    (~s, ~t, ~\Delta u, ~\Delta v)_2\\
    ... \\
    (~s, ~t, ~\Delta u, ~\Delta v)_{M}
  \end{array}
\right]
}_{\mathbf{A}}
\underbrace{\left[
  \begin{array}{c}
	n_1\\ n_2\\ n_3\\ n_4
  \end{array}
\right]
}_{\vec{\mathbf{n}}}
= \mathbf{0}.
\end{equation}
We then use singular value decomposition to calculate $\mathbf{A}^\top \mathbf{A}=\mathbf{U}\mathbf{D}\mathbf{U}^\top$, and the linear least-squares solution to $\vec{\mathbf{n}}$ is the column of $\mathbf{U}$ associated with the smallest eigenvalue in $\mathbf{D}$, where the smallest eigenvalue is the least-squares error $E(u,v)$. Smaller errors imply better linearity, and larger errors indicate that the feature deviates strongly from the hyperplane. Because this error $E(u,v)$ describes the linearity of the LFD feature, we call this the LF-linearity. This important property is used to define the regional term in the energy function. Figure \ref{fig:LF-lin} shows an example of the visualized LF-linearity.

\subsection{Occlusion detector} \label{sec:Occ_detector}
\begin{figure}[t]
	\centering
	\includegraphics[width=.9\columnwidth]{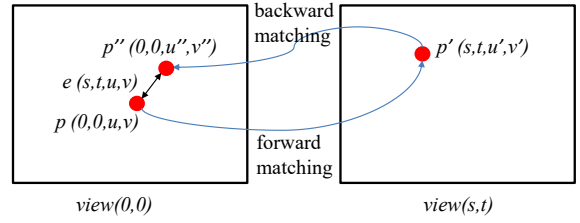}
	\caption{Checking the consistency of the forward and backward matching between a pair of viewpoints.}
    \label{fig:forward-backword_matching}
\end{figure}

The background can be occluded by foreground objects in different viewpoints. This is an important cue for determining the boundaries between the foreground and background. The occlusion boundary is often detected by comparing the appearance of points over time as the camera or object moves. In a light-field image, we detect occlusion points by checking the consistency of the forward and backward matching between a pair of viewpoints, as illustrated in Fig. \ref{fig:forward-backword_matching}.

We denote an arbitrary point in the image captured by the central viewpoint $view(0,0)$ as $p(0,0,u,v)$, and the corresponding point in the image captured by another viewpoint $view(s,t)$ as $p'(s, t, u',v')$.
Here, $(s,t)$ are the coordinates of the viewpoint $view(s,t)$, and $(u,v)$ are the coordinates of the point in the image plane (as shown in Fig. \ref{fig:LF_propagation}).
We also attempt to find the point in the central viewpoint $view(0,0)$ that corresponds to $p'(s, t, u',v')$, which we denote as $p''(0, 0, u'',v'')$.

The consistency is independent of the intensity at each point, so we can simply calculate the geometric error of the forward and backward matching:
\begin{equation}
e(s,t,u,v)=dist(p(0,0,u,v),p''(0, 0, u'',v'')),
\label{eq:consistency}
\end{equation}
where $dist(p,p'')$ is the Euclidean distance between $p$ and  $p''$.

In the non-occlusion case, points $p(0,0,u,v)$  and $p''(0, 0, u'',v'')$ should be very close, which means the error $e(s,t,u,v)$ will be very small.
If this consistency requirement is not satisfied, the point is either occluded in the corresponding viewpoint, or the optical flow has been incorrectly estimated.
The small values are mainly from noise, and the large error values do not have much physical meaning. Hence, we define the LF-consistency $c(s,t,u,v)$ by binarizing the error $e(s,t,u,v)$.
\begin{equation}
c(s,t,u,v)=
\begin{cases}
0, ~~~~~~~~~~~e(s,t,u,v)< \tau\\
1, ~~~~~~~~~~~e(s,t,u,v)\geq \tau
\end{cases}
.
\label{eq:occ}
\end{equation}
where $\tau$ is a tolerance interval that allows the noise introduced by the optical flow calculation.
We assign zeros to consistent points and ones to inconsistent points.

\begin{figure}[t]
	\centering
	\includegraphics[width=0.9\columnwidth]{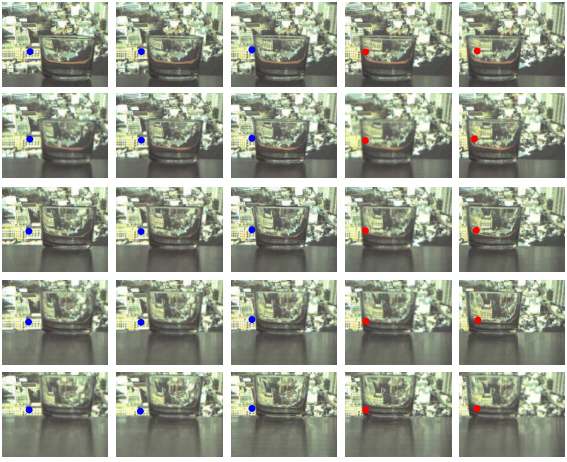}
	\caption{An example of the pixel at occlusion boundary. The pixel in the center viewpoint can find the corresponding point from the viewpoints in the left 3 columns (shown in blue dots), but the corresponding point cannot be found in the right viewpoints where the point is occluded by the foreground object (shown in red dots). The blue dots have good LF-consistency, while the red dots are with poor LF-consistency.}
    \label{fig:occ_example}
\end{figure}
\begin{figure}[t]
	\begin{center}
	\subfigure[$\theta=0$]{
		\includegraphics[width=0.22\columnwidth]{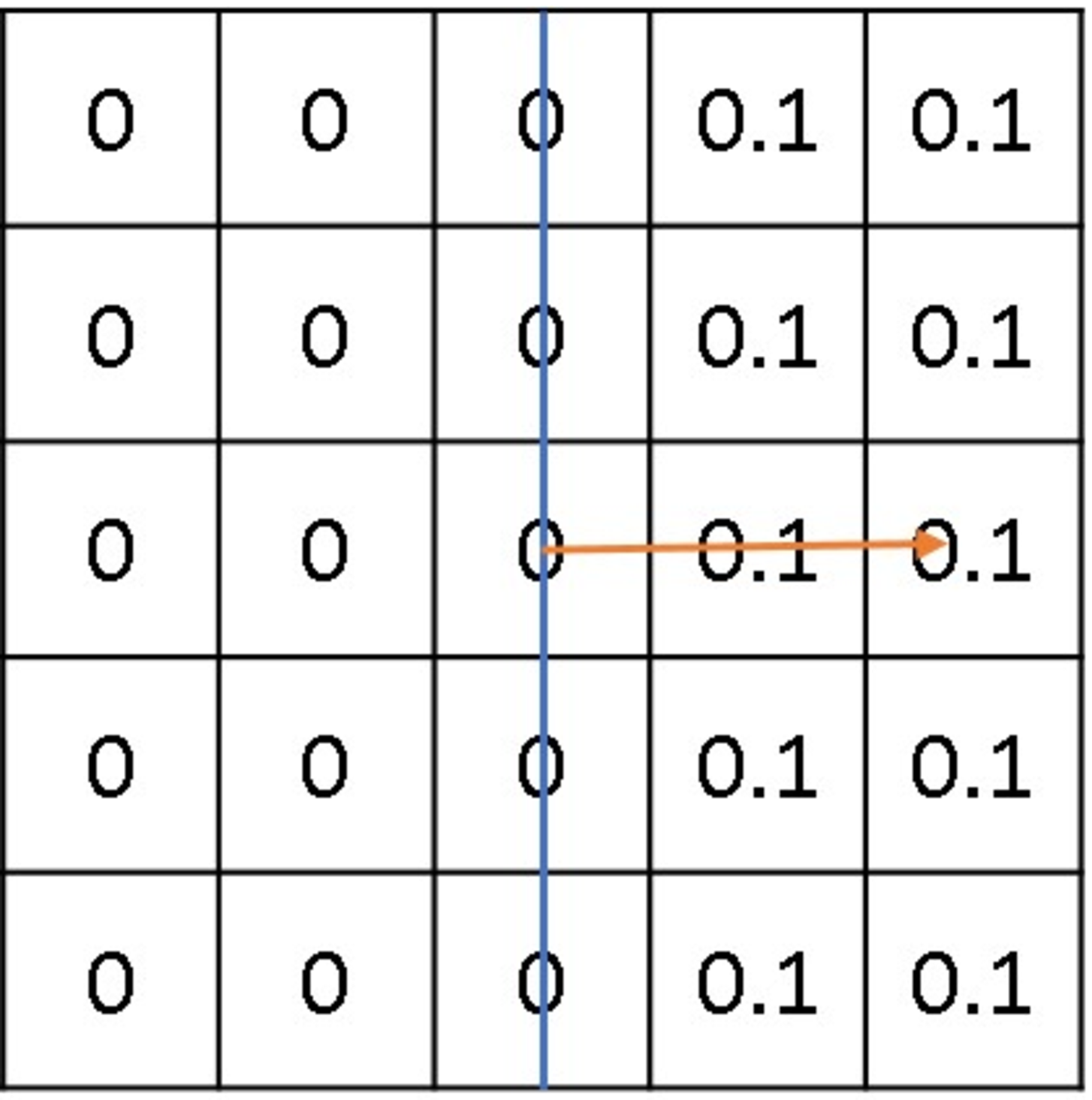}
		\label{fig:f0}
	}
	\subfigure[$\theta={45}$]{
		\includegraphics[width=0.22\columnwidth]{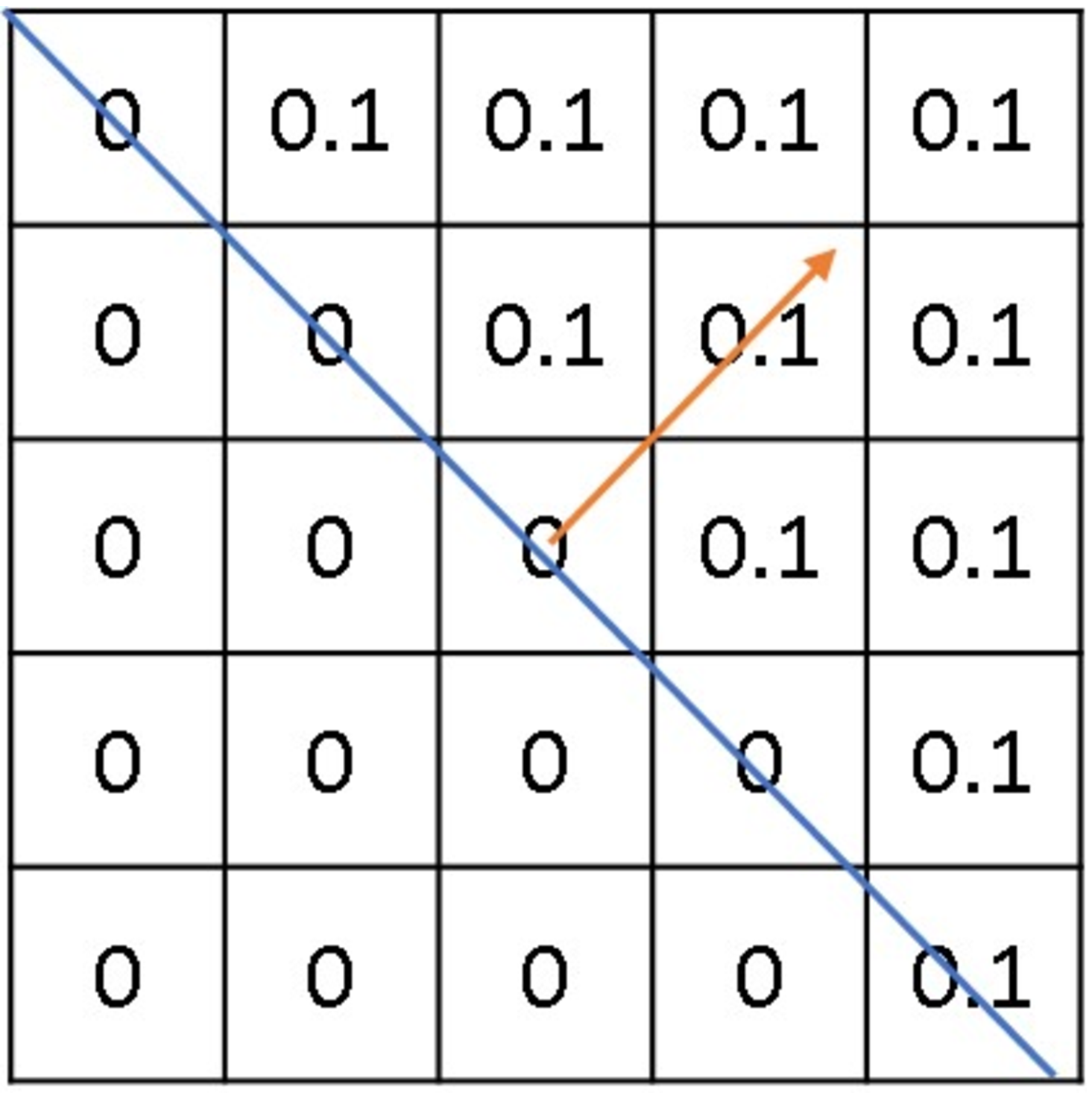}
		\label{fig:f45}
	}
	\subfigure[$\theta={90}$]{
		\includegraphics[width=0.22\columnwidth]{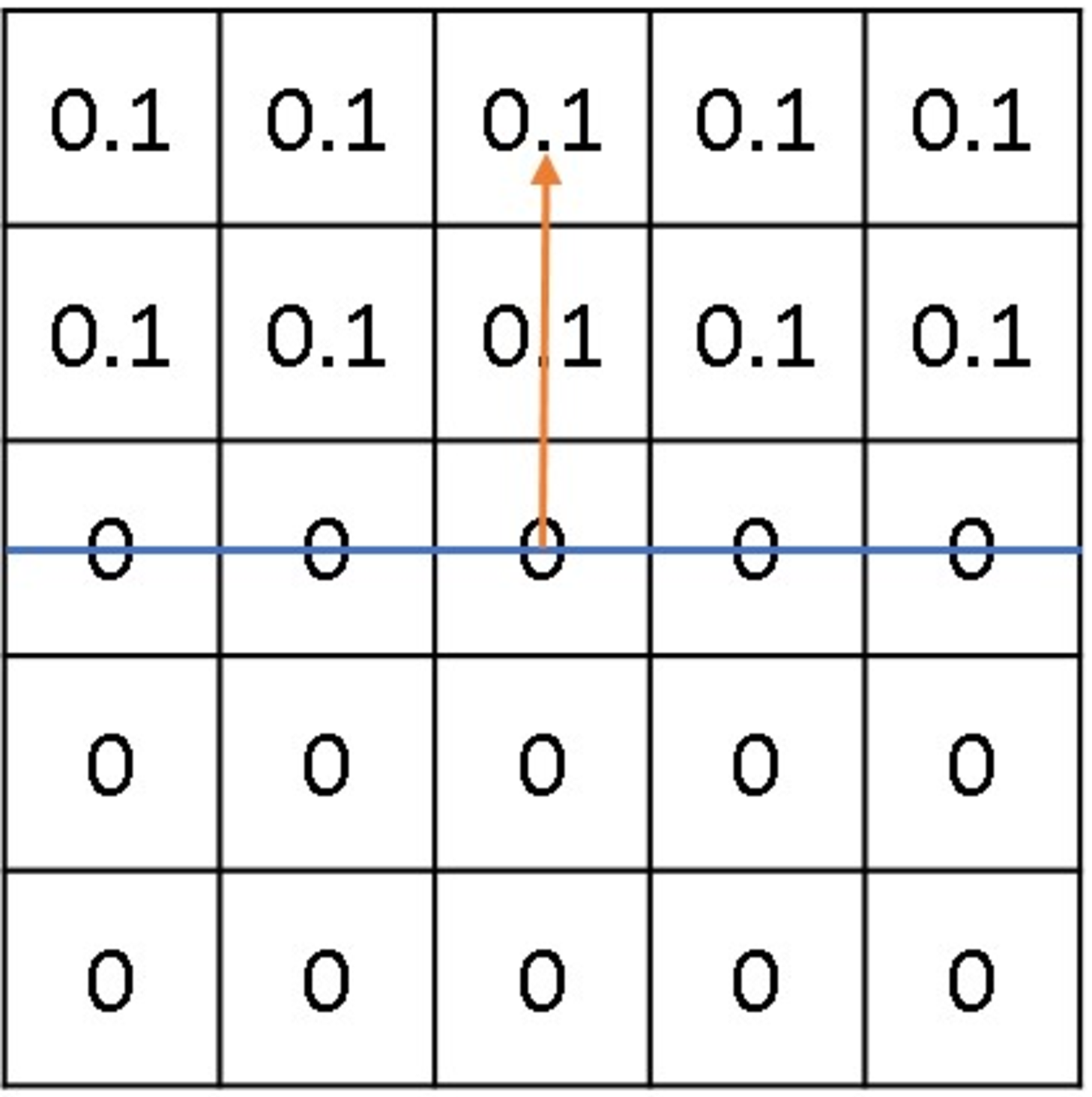}
		\label{fig:f90}
	}
    \subfigure[$\theta={135}$]{
		\includegraphics[width=0.22\columnwidth]{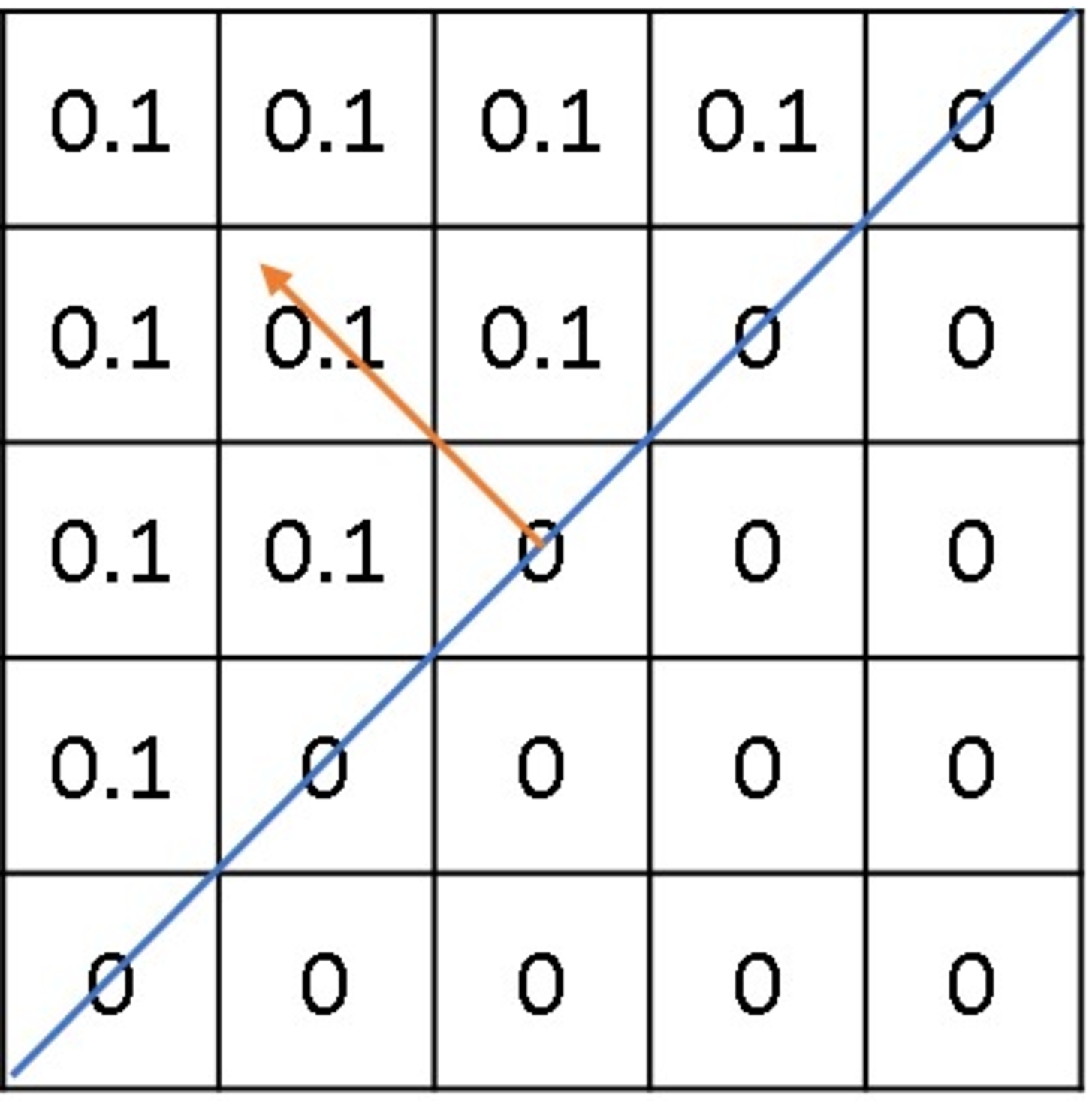}
		\label{fig:f135}
	}
    \subfigure[$\theta={180}$]{
		\includegraphics[width=0.22\columnwidth]{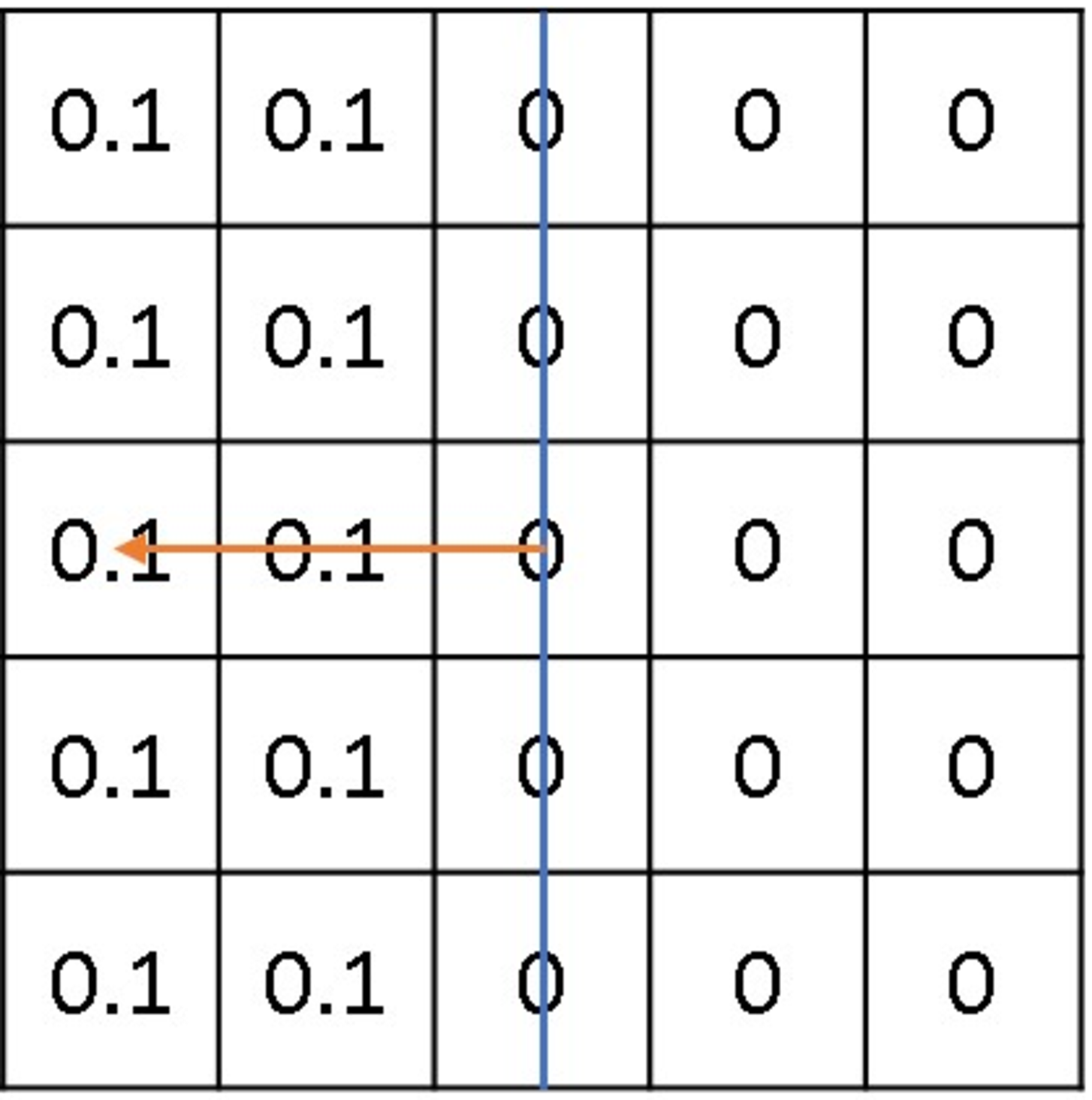}
		\label{fig:f180}
	}
    \subfigure[$\theta={225}$]{
		\includegraphics[width=0.22\columnwidth]{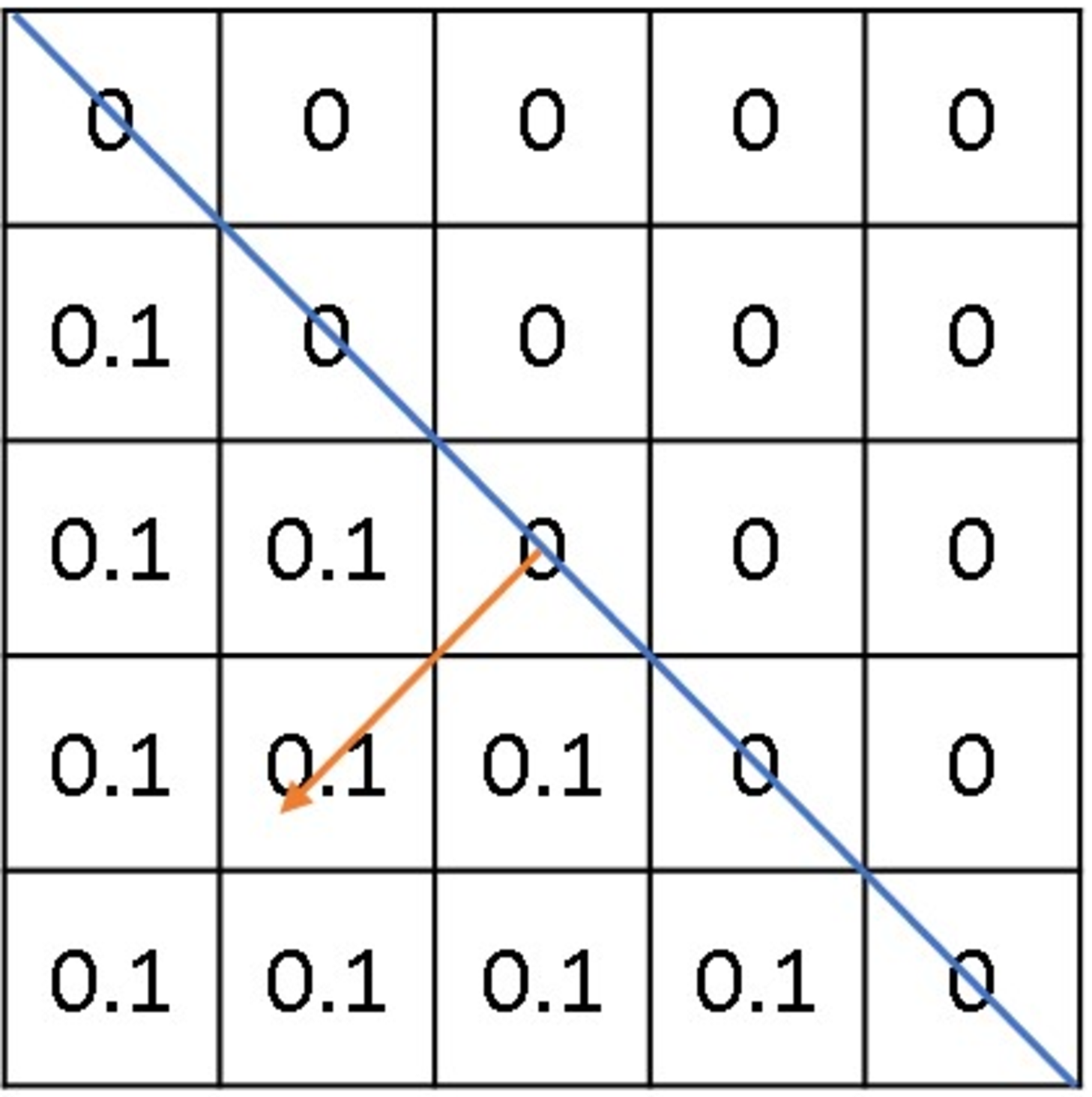}
		\label{fig:f225}
	}
    \subfigure[$\theta={270}$]{
		\includegraphics[width=0.22\columnwidth]{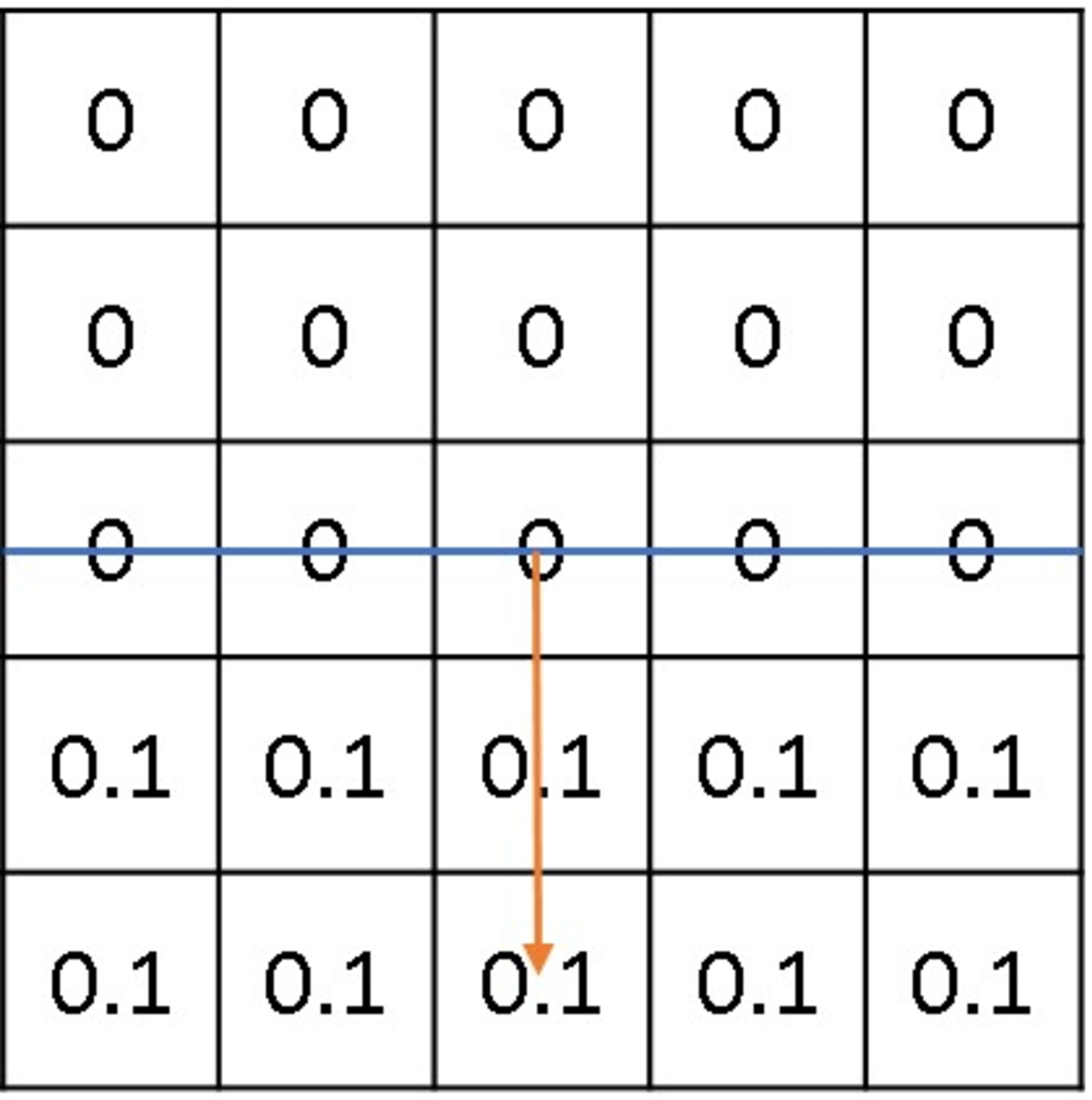}
		\label{fig:f270}
	}
    \subfigure[$\theta={315}$]{
		\includegraphics[width=0.22\columnwidth]{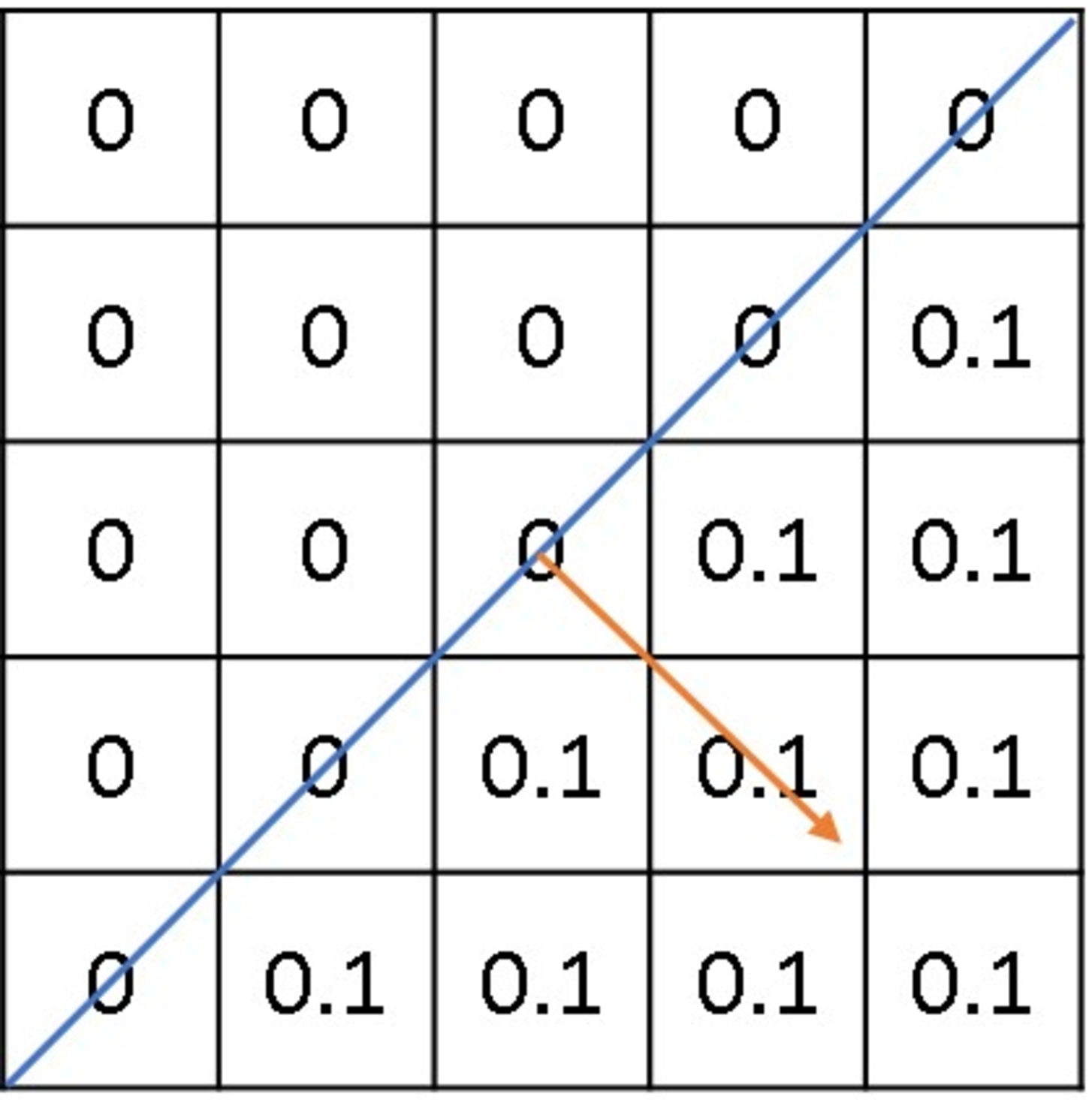}
		\label{fig:f315}
	}
	\caption{Occlusion detectors $F(s,t,\theta)$ in 8 different directions.}
	\label{fig:occ_detector}
\end{center}
\end{figure}
\begin{figure}[t]
	\begin{center}
	\subfigure[Central view]{
		\includegraphics[width=0.3\columnwidth]{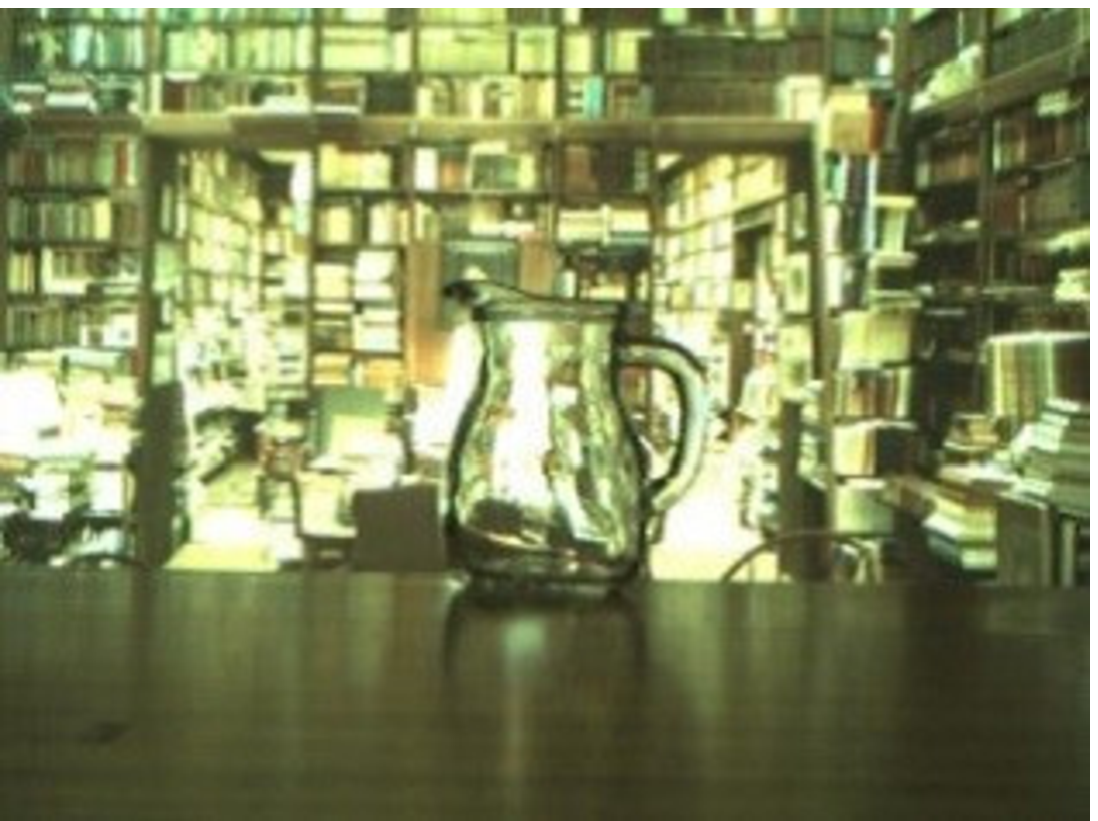}
		\label{fig:center_inter}
	}
	\subfigure[LF-linearity]{
		\includegraphics[width=0.3\columnwidth]{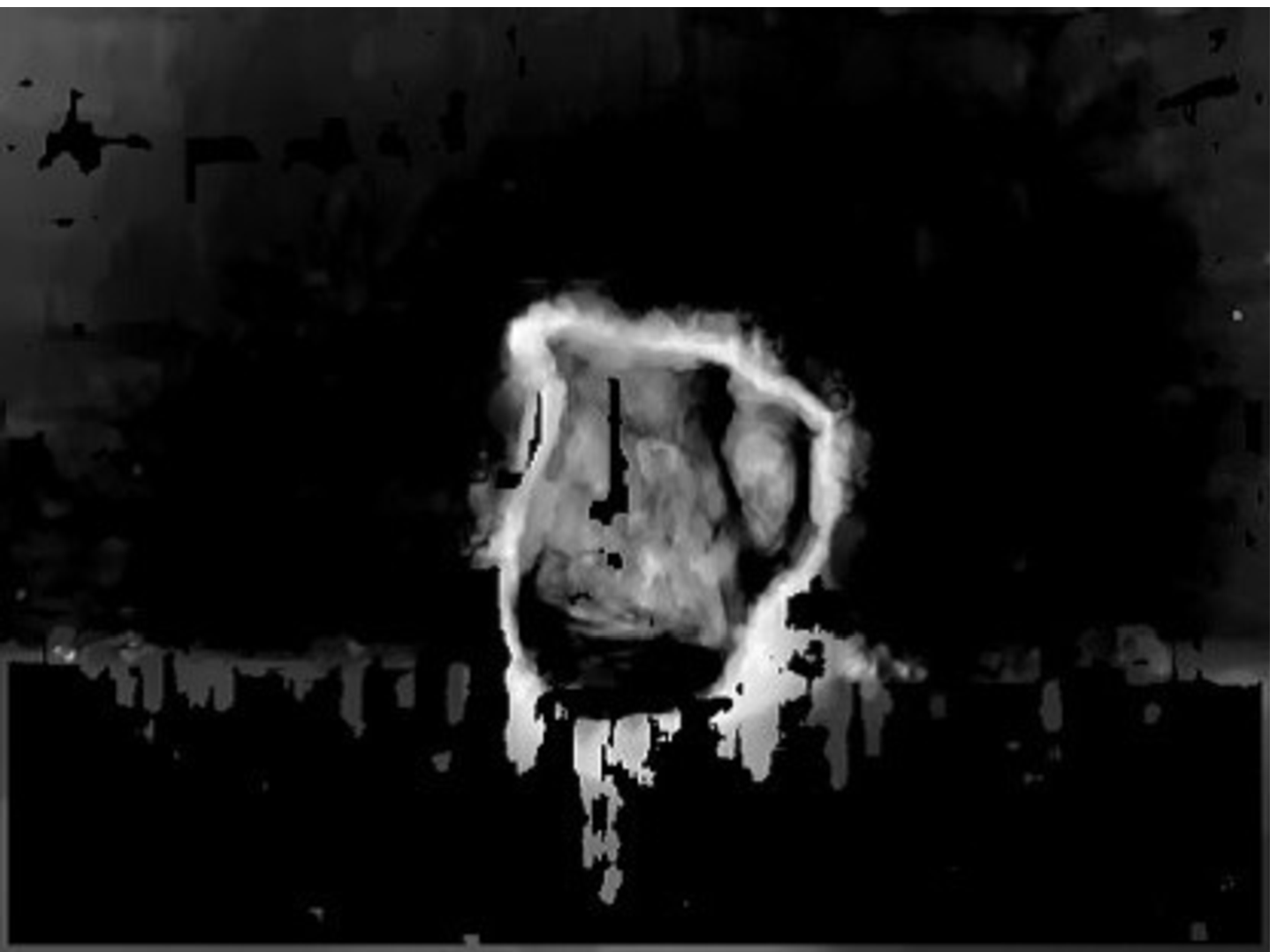}
		\label{fig:LF-lin}
	}
    \subfigure[Detected occlusion]{
		\includegraphics[width=0.3\columnwidth]{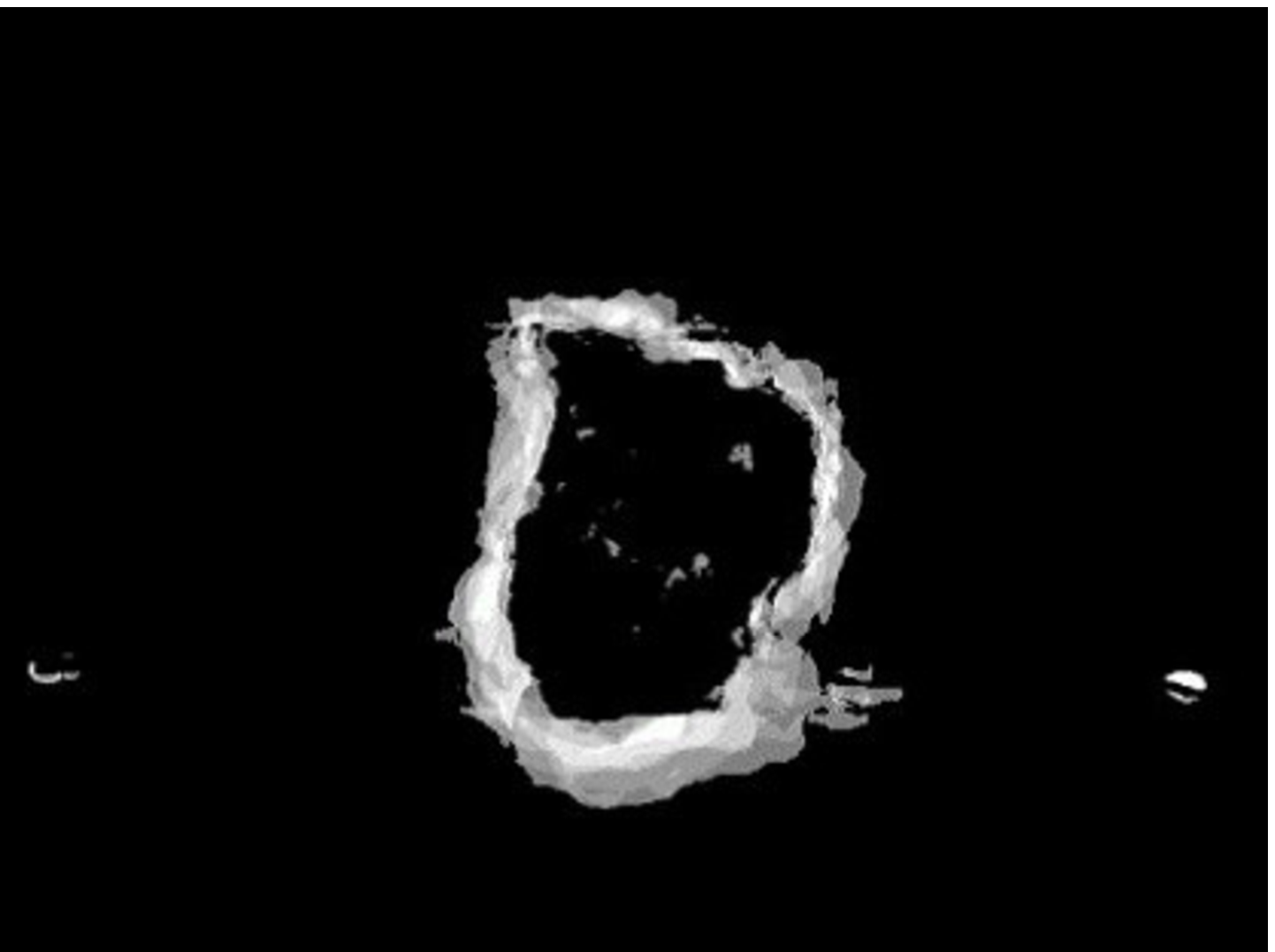}
		\label{fig:occ_response}
	}
 	\caption{An example of visualized the LF-linearity and detected occlusion.}
	\label{fig:inter_result}
\end{center}
\end{figure}
The LF-consistency has different patterns when the occlusion boundary appears in different directions. Fig. \ref{fig:occ_example} shows an example of a point that has both consistency and inconsistency in different viewpoints.
Based on our observations, we have designed a series of occlusion detectors $F(s,t,\theta)$ to detect the occlusion boundaries between foreground and background. The detectors of $5 \times 5$ case, which are used in our experiments, are shown in Fig. \ref{fig:occ_detector}, and $\theta$ is the normal direction of the occlusion boundary.
The size of occlusion detector is corresponding to the number of viewpoints.
The non-zero values in the detector indicate a point is occluded in the corresponding viewpoint.

We use $c(s,t,u,v)$ and $F(s,t,\theta)$ to decide the likelihood of a pixel $(u,v)$  being the occlusion boundary in the direction $\theta$:
\begin{equation}
O(u,v, \theta)=
\sum_s \sum_t c(s,t,u,v)\cdot F(s,t,\theta)
.
\label{eq:occ_det}
\end{equation}

The direction with largest response of all the detectors will be chosen as the occlusion direction:
\begin{equation}
\tilde{\theta}(u,v)=\argmax_{\theta}O(u,v,\theta)
.
\label{eq:occ_det2}
\end{equation}
An example of the detected occlusion is shown in Fig. \ref{fig:occ_response}.
\section{TransCut: graph-cut segmentation for transparent object} \label{sec:energy_func}
The goal of this work is to segment transparent objects by using LF-linearity and occlusion detector.
We formulate the segmentation task as a pixel labeling problem with two labels (transparent objects as the foreground and other objects as the background). Later part of this paper, we describe each pixel as $p = (0,0,u, v)$ and some variables with subscript $p$ indicate the variables at pixel $p$ of the center viewpoint, since we solve the pixel labeling problem in 2D image space.
Similar to other segmentation methods \cite{boykov2006graph, rother2004grabcut}, we define an energy function to evaluate the labeling problem:
\begin{equation}
E(l)= \sum _{p\in P}R_p(l_p)+ \alpha \sum _{(p,q)\in N}B_{p,q}\cdot \delta (l_p, l_q),
\label{eq:energy}
\end{equation}
where  $l_p$ is the label of an image pixel $p$ ($l_p=0$ denotes a background pixel, $l_p=1$ denotes a foreground pixel), $R_p(l_p)$ is the regional term that measures the penalties for assigning $l_p$ to $p$, $B_{p,q}$ is the boundary term for measuring the interaction potential between pixels $p$ and $q$, $N$ is the neighborhood set, $\alpha$ adjusts the balance between $R_p(l_p)$ and $B_{p,q}\cdot \delta (l_p, l_q)$, and
\begin{equation}
\delta (l_p, l_q)=
\left\{
\begin{aligned} 
1, & ~~ if~~ l_p\neq l_q\\
0, & ~~ if~~ l_p=l_q
\end{aligned}
\right.
\end{equation}
The segmentation task aims to determine the labeling that minimizes Eq. \ref{eq:energy}. We use the graph-cut method to optimize the energy function.

\begin{figure}[t]
	\centering
	\includegraphics[width=0.7\columnwidth]{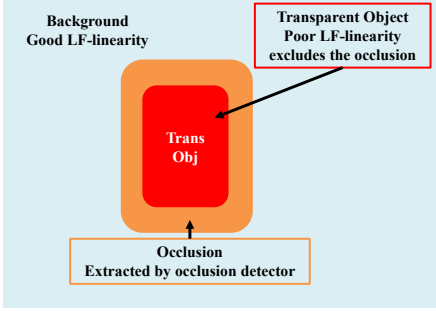}
	\caption{Properties of different components in an image containing a transparent object. The Lambertian background (blue) has good LF-consistency, the transparent object (red) has poor LF-linearity excludes the occlusion area, and the occlusion boundary (orange) can be detected by occlusion detector.}
    \label{fig:layer_property}
\end{figure}
\subsection{Regional term} \label{sec:regional}
We assume that all Lambertian objects in the image should be labeled as background, and the refractive transparent object should be labeled as the foreground. As illustrated in Fig. \ref{fig:layer_property}, the background and the occluded areas (shown in blue and orange) should be labeled as the background, and the transparent object (red) should be labeled as foreground.

The Lambertian object has good LF-linearity while the transparent object has poor LF-linearity. The occlusion area also has poor LF-linearity and can be detected by the occlusion detector, so the transparent object locates in area with poor LF-linearity other than the occlusion area.
The case of the occlusion area with good LF-linearity rarely occurs because, when the forward-backward matching is not consistent, the LF-linearity will be poor. Therefore, the region with good LF-linearity should be background.
When a pixel belongs to the background, the penalty for labeling this pixel as a Lambertian object or occlusion area should be low, while the penalty for labeling this pixel as part of a transparent object should be high. The opposite is true when a pixel belongs to the foreground.

Before defining the regional term of the energy function, we first scale the LF-linearity $E(u,v)$ to the range $[0,1]$ using a sigmoid function:
\begin{equation}\label{eq:scale_E}
\tilde{E}_p  = sigmoid(E(u,v),a,b),
\end{equation}
where $sigmoid(\varphi,a,b)$ is the function:
\begin{equation}\label{eq:sigmoid}
sigmoid(\varphi,a,b)=\frac{1}{1+exp(-a (\varphi-b))},
\end{equation}
$a$ controls the steepness of the function, and $b$ is the shift, which acts as the threshold value here.

The regional term for a pixel $p$ is defined as:
\begin{equation}\label{eq:R_p(0)}
R_p(0)  = \beta \tilde{E}_p\cdot(1-\tilde{O}_p),
\end{equation}
\begin{equation}\label{eq:R_p(1)}
R_p(1) =  \tilde{E}_p\cdot \tilde{O}_p + (1-\tilde{E}_p),
\end{equation}
where $\tilde{O}_p = O(u,v,\tilde{\theta})$,
which is the maximum response from the occlusion detectors descried in Eq. \ref{eq:occ_det} and Eq. \ref{eq:occ_det2}.
$R_p(0)$ assigns a large penalty to pixels that have poor LF-linearity exclude the occlusion area,
and $R_p(1)$ assigns a large penalty to pixels with poor LF-linearity inside the occlusion area or pixels with good LF-linearity. $\beta$ adjusts the balance between $R_p(0)$ and $R_p(1)$.
\subsection{Boundary term} \label{sec:smooth}
In the boundary term of the energy function, we must define the pairwise potentials between two neighboring pixels. We use the 4-neighbor system, so each pixel has two horizontal neighboring pixels and two vertical neighboring pixels.
We utilize the maximum response of the occlusion detectors to assign pairwise potentials.

\begin{figure}[t]
	\centering
	\includegraphics[width=.5\columnwidth]{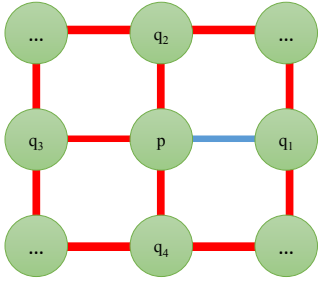}
	\caption{Definition of energy for the pairwise potential $B_{p,q}$. The example shows the maximum response $O_{p,\tilde{\theta}}$ comes from $\tilde{\theta}=0$, hence we assign a small penalty $B_{p,q_1}$ to the corresponding edge (blue)}
    \label{fig:image_grid}
\end{figure}

The boundary term applies a penalty when neighboring pixels $p$, $q$ are assigned different labels.
Given a pixel $p$ (see Fig. \ref{fig:image_grid}), the weight of its 4 neighboring edges can be described as:
\begin{align}
&\begin{cases}
w_{p,q_1} = \tilde{O}_p \\
w_{p,q_2} = w_{p,q_3} =w_{p,q_4} = 0
\end{cases}
, \text{if} ~~\tilde{\theta}=0, \label{eq:bd_weight} \\
&\begin{cases}
w_{p,q_1} = w_{p,q_2} = \tilde{O}_p/\sqrt{2} \\
w_{p,q_3} = w_{p,q_4} = 0
\end{cases}
, ~~~\text{if}~~\tilde{\theta}=45,
\label{eq:bd_weight}
\end{align}
and so forth.
The weight for each edge is calculated twice as $w_{p,q}$ and $w_{q,p}$, and the penalty for assigning different labels to $p$ and $q$ is defined as:
\begin{equation}
B_{p,q}= exp(-\gamma \cdot (w_{p,q}+w_{q,p})).
\label{eq:bd_term}
\end{equation}
The weight is small in the background and foreground regions. The penalty of the region is high in the case of assigning different labels to the neighboring pixels. It works easy to propagate the same labels in the same regions. In contrast, the occlusion boundary will have large values of $\tilde{O}_p$, and it stop to propagate the label between the different regions. $\gamma$ controls the rate of the importance of the penalty.
\section{Experiments}\label{sec:experiment}
As there are no light field datasets available for the evaluation of transparent object segmentation, we captured the necessary data ourselves.
We shall demonstrate our proposed transparent object segmentation method on various examples, including single and multiple objects segmentation with different backgrounds, and compare with other methods such as finding glass \cite{findingGlass}.
\subsection{Assumptions}
To ensure the effectiveness of the matching process, our experiments were conducted under the following assumptions:
\begin{itemize}
	\item All viewpoints of the light-field camera can capture the entirety of the target objects.
    \item The degree of reflection on the surface of the target objects is relatively low.
	\item Background is relatively far away with rich texture.
\end{itemize}
\subsection{Results and discussion}
\begin{figure}[t]
	\begin{center}
	\subfigure[Object 1]{
		\includegraphics[width=0.22\columnwidth]{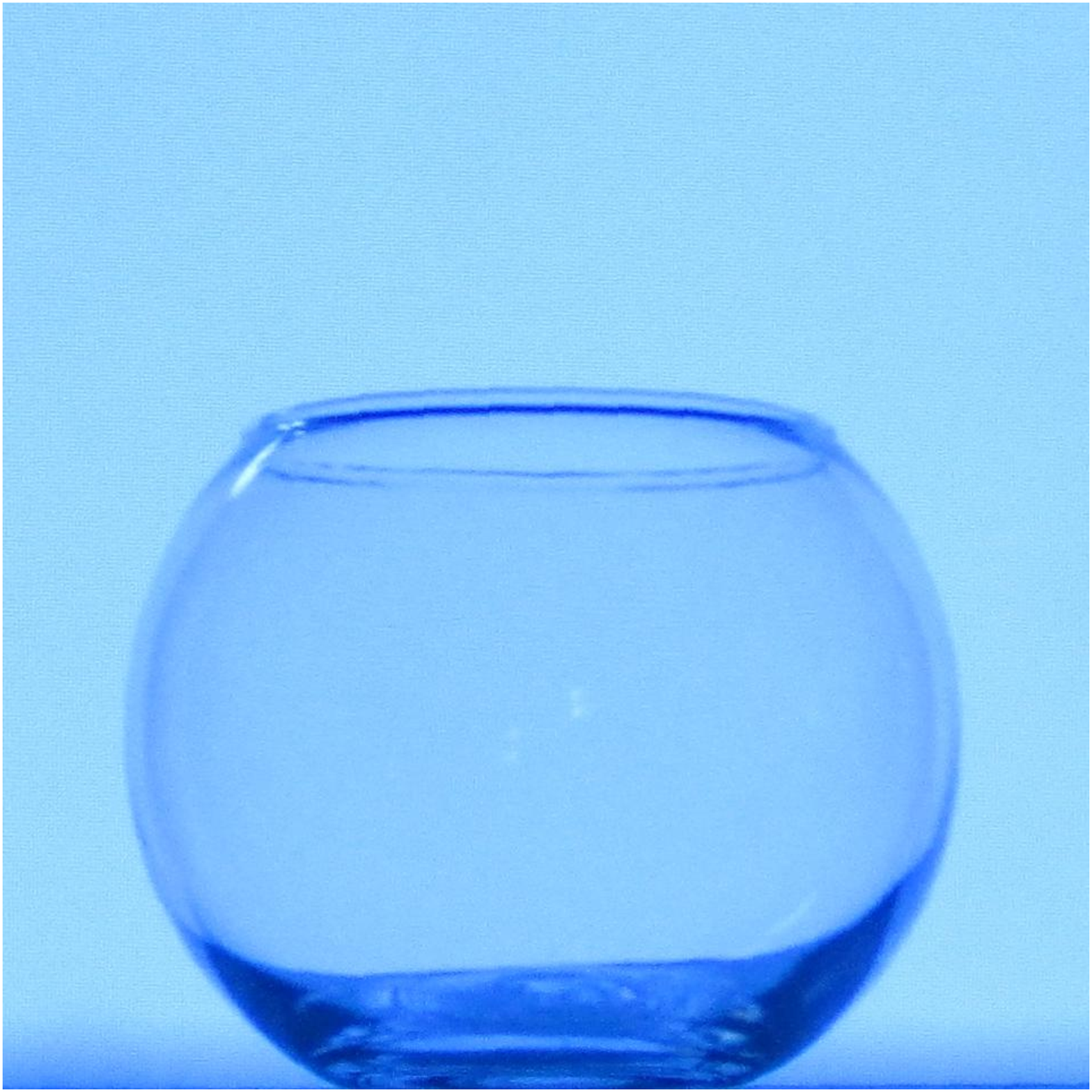}
		\label{fig:obj1}
	}
	\subfigure[Object 2]{
		\includegraphics[width=0.22\columnwidth]{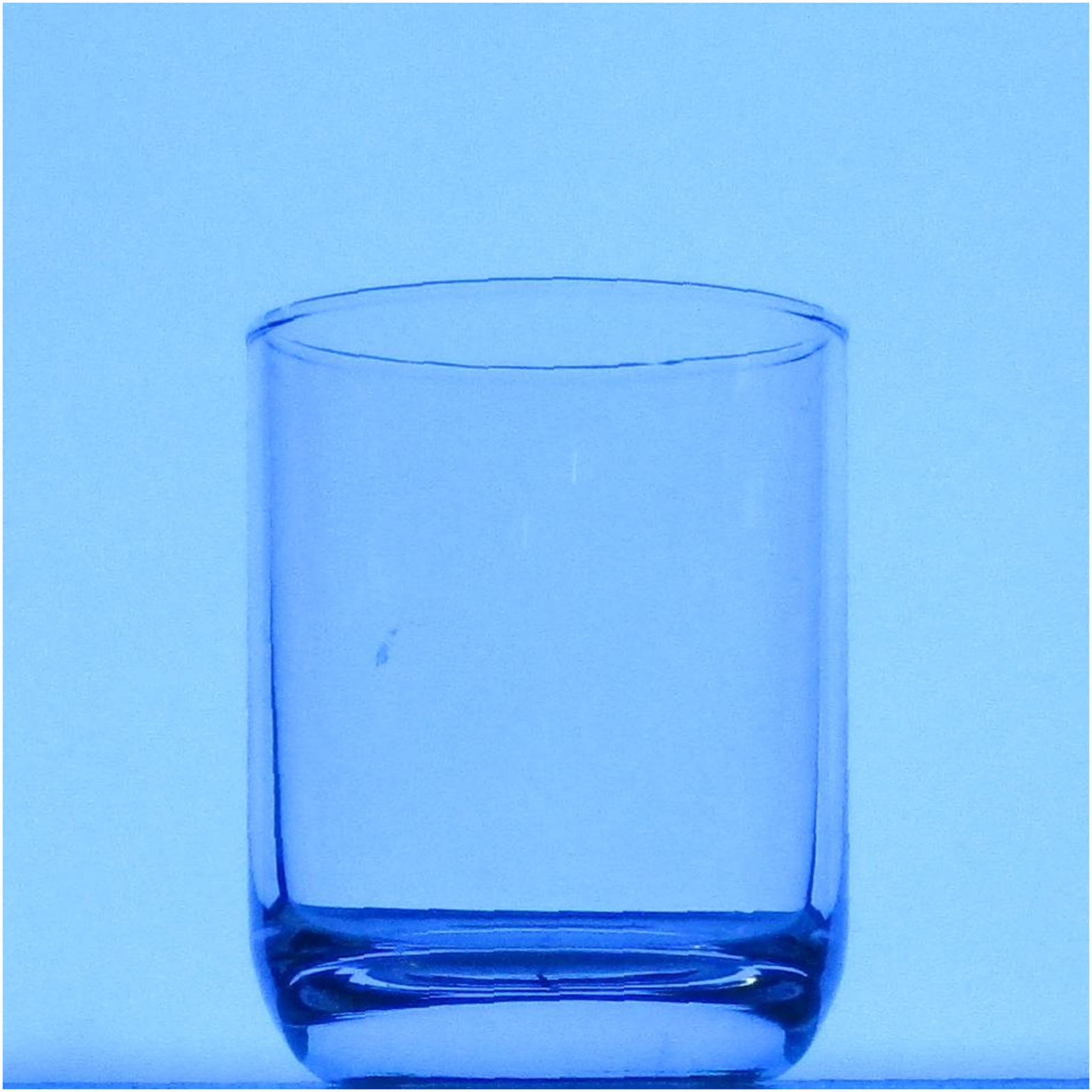}
		\label{fig:obj2}
	}
	\subfigure[Object 3]{
		\includegraphics[width=0.22\columnwidth]{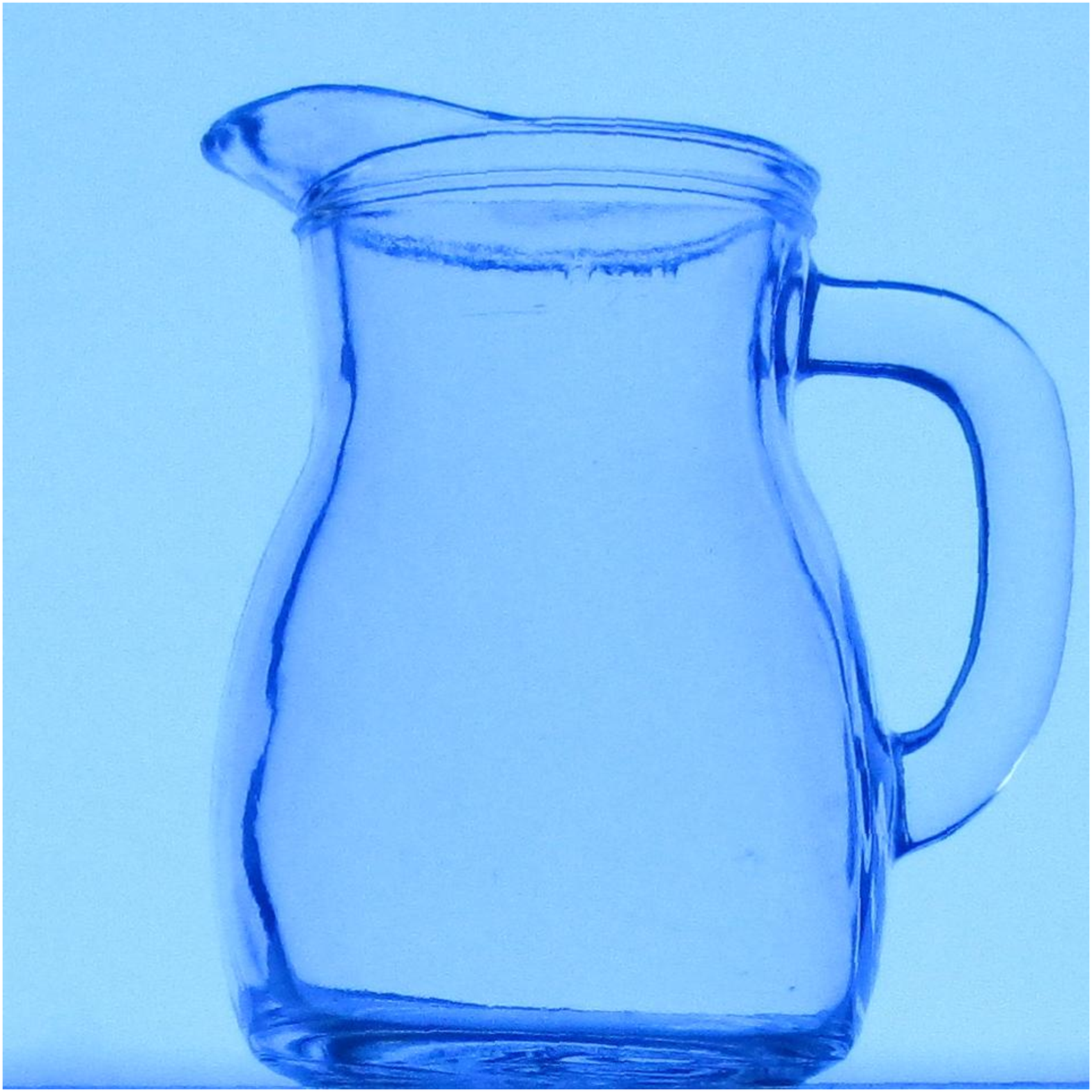}
		\label{fig:obj3}
	}
    \subfigure[Object 4]{
		\includegraphics[width=0.22\columnwidth]{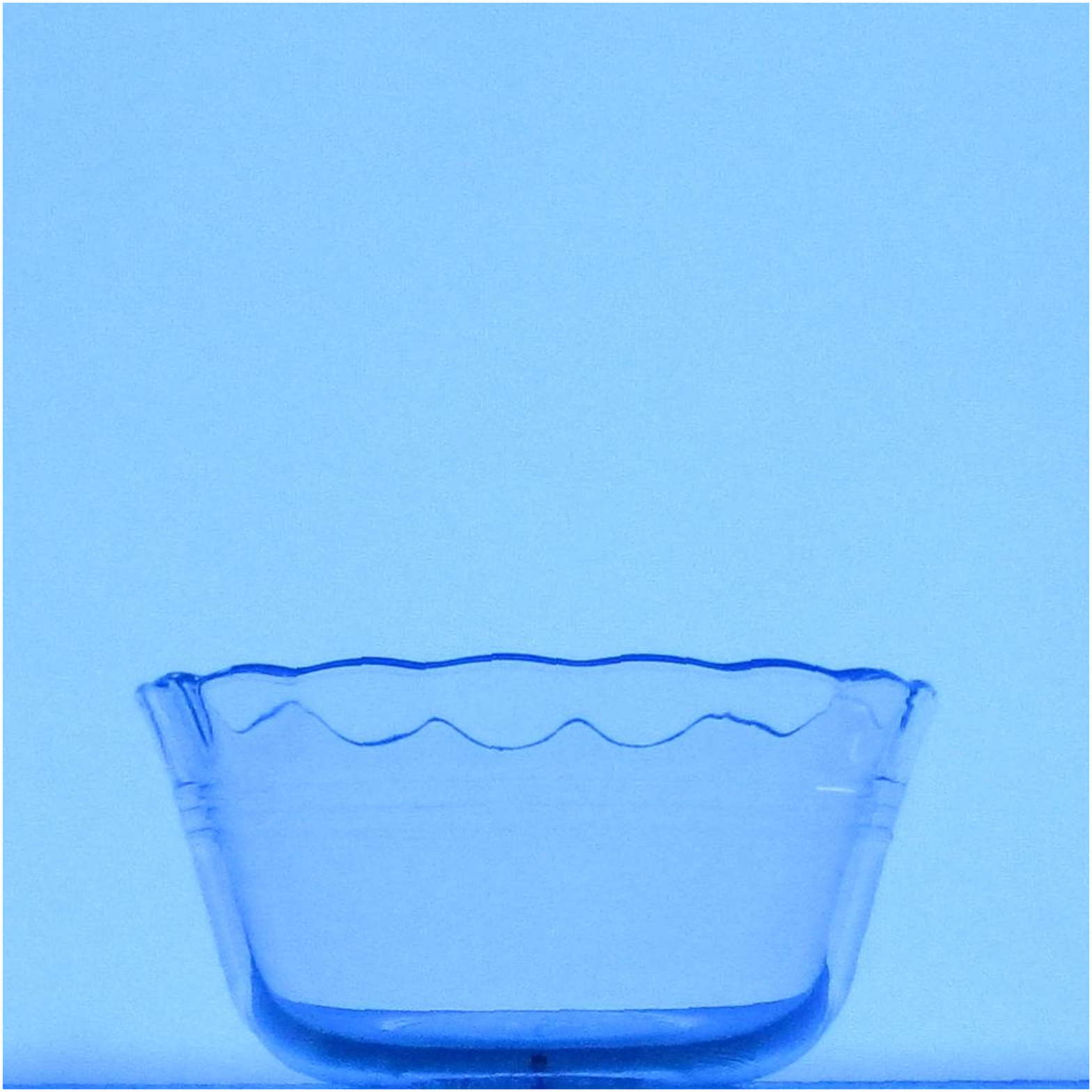}
		\label{fig:obj4}
	}
    \subfigure[Object 5]{
		\includegraphics[width=0.22\columnwidth]{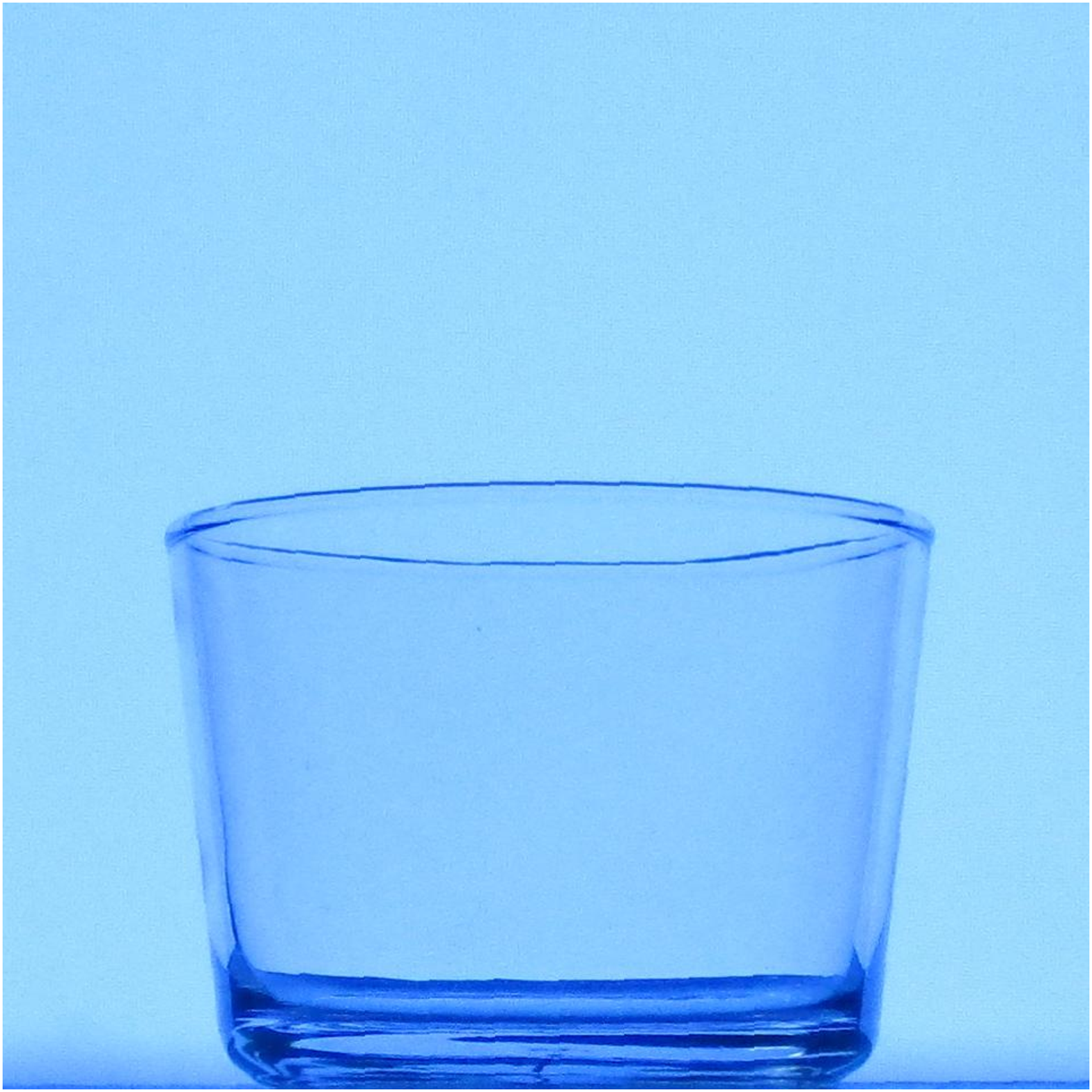}
		\label{fig:obj5}
	}
    \subfigure[Object 6]{
		\includegraphics[width=0.22\columnwidth]{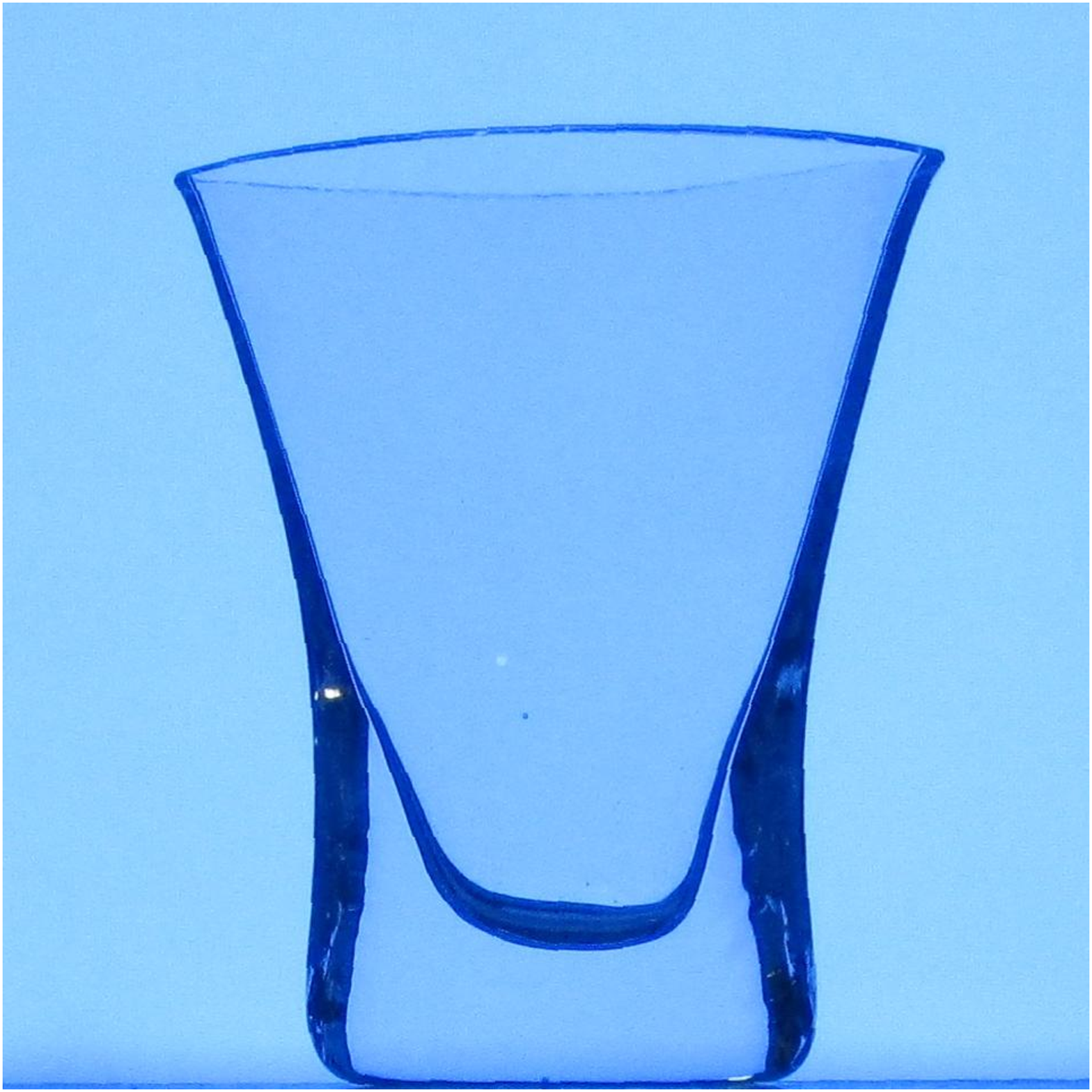}
		\label{fig:obj6}
	}
    \subfigure[Object 7]{
		\includegraphics[width=0.22\columnwidth]{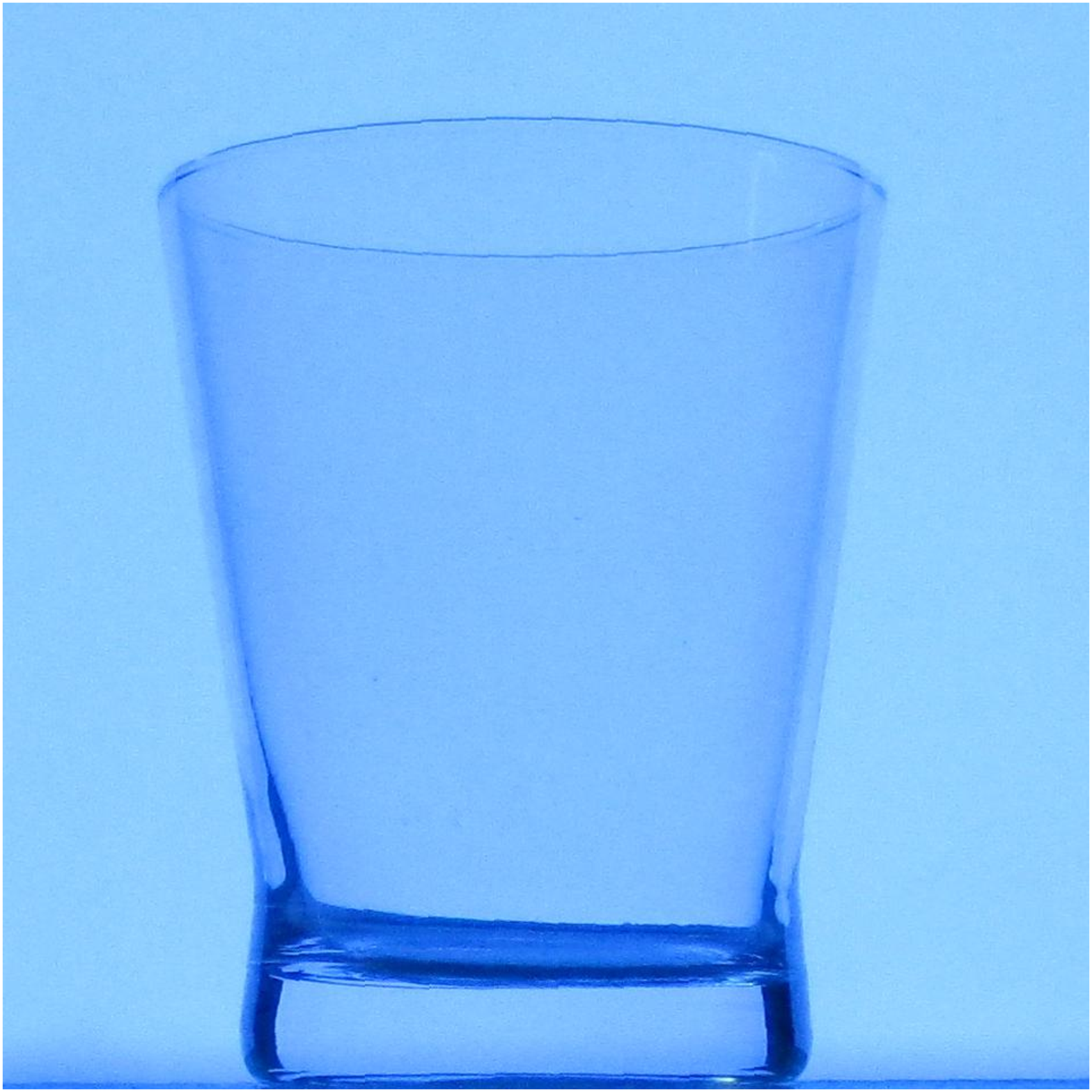}
		\label{fig:obj7}
	}
	\caption{seven transparent objects of various shapes for the experiments.}
	\label{fig:obj}
\end{center}
\end{figure}

\begin{figure}[t]
	\begin{center}
	\subfigure[Scene 1]{
		\includegraphics[width=0.22\columnwidth]{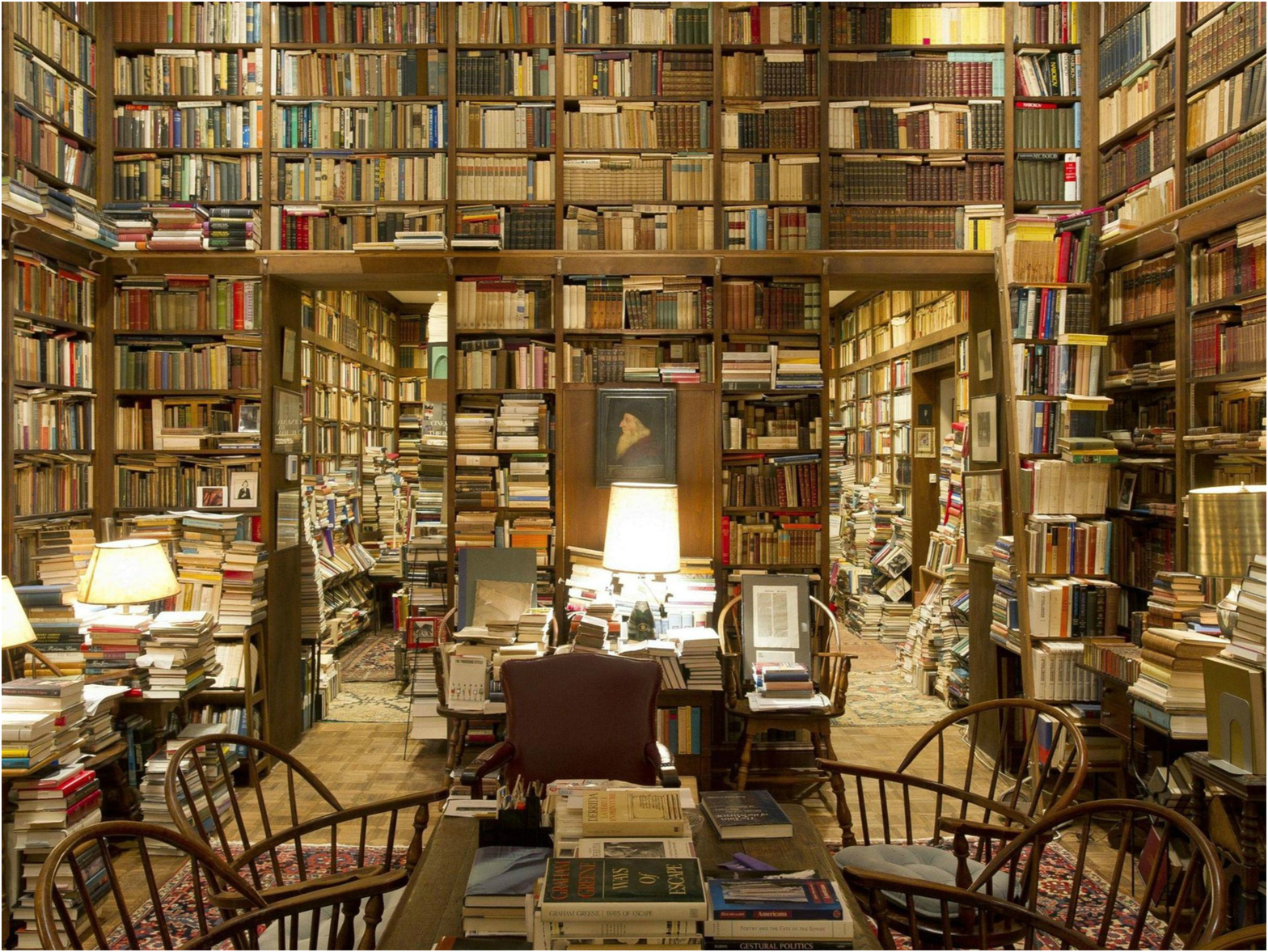}
		\label{fig:scn4}
	}
	\subfigure[Scene 2]{
		\includegraphics[width=0.22\columnwidth]{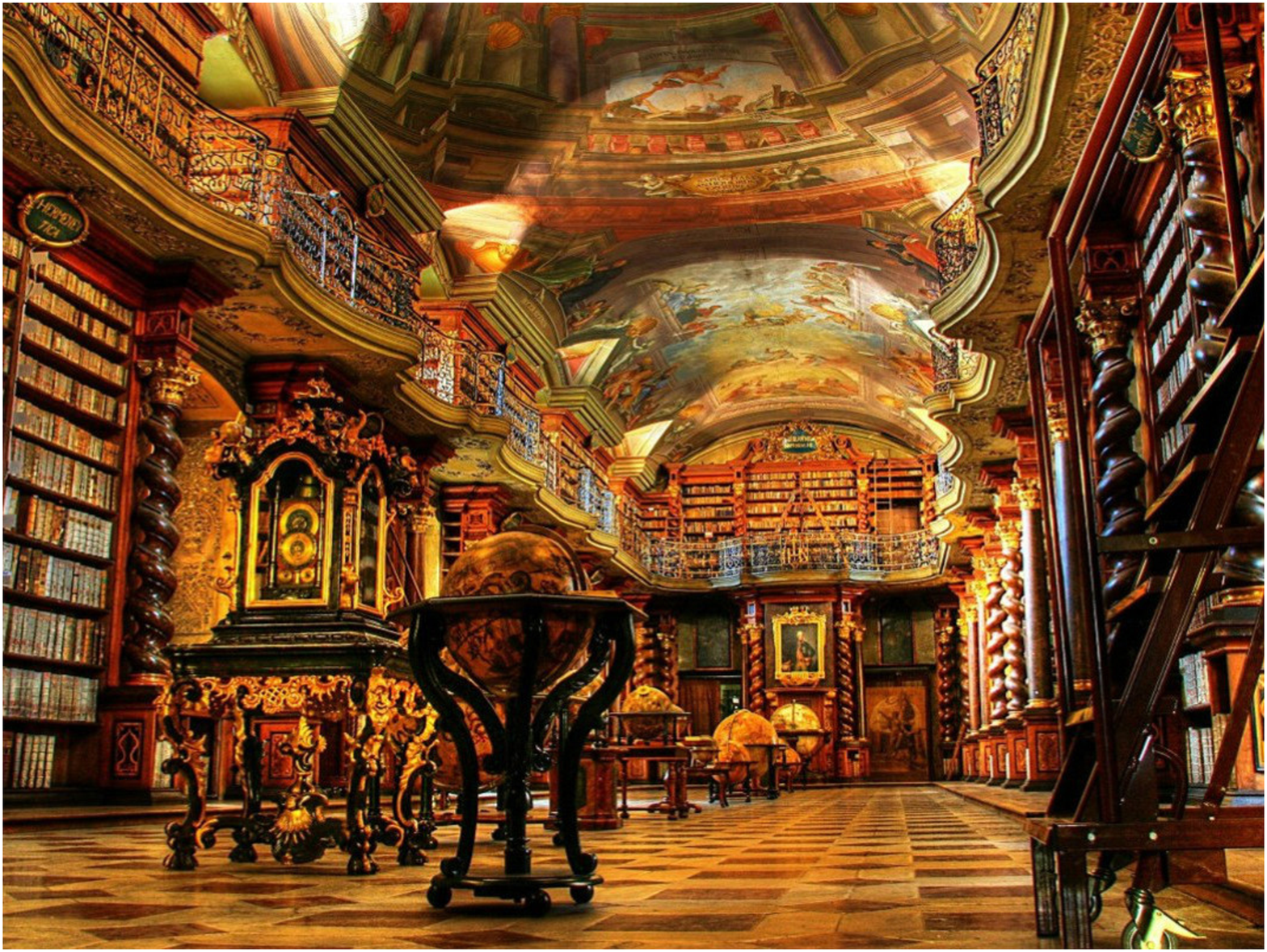}
		\label{fig:scn2}
	}
    \subfigure[Scene 3]{
		\includegraphics[width=0.22\columnwidth]{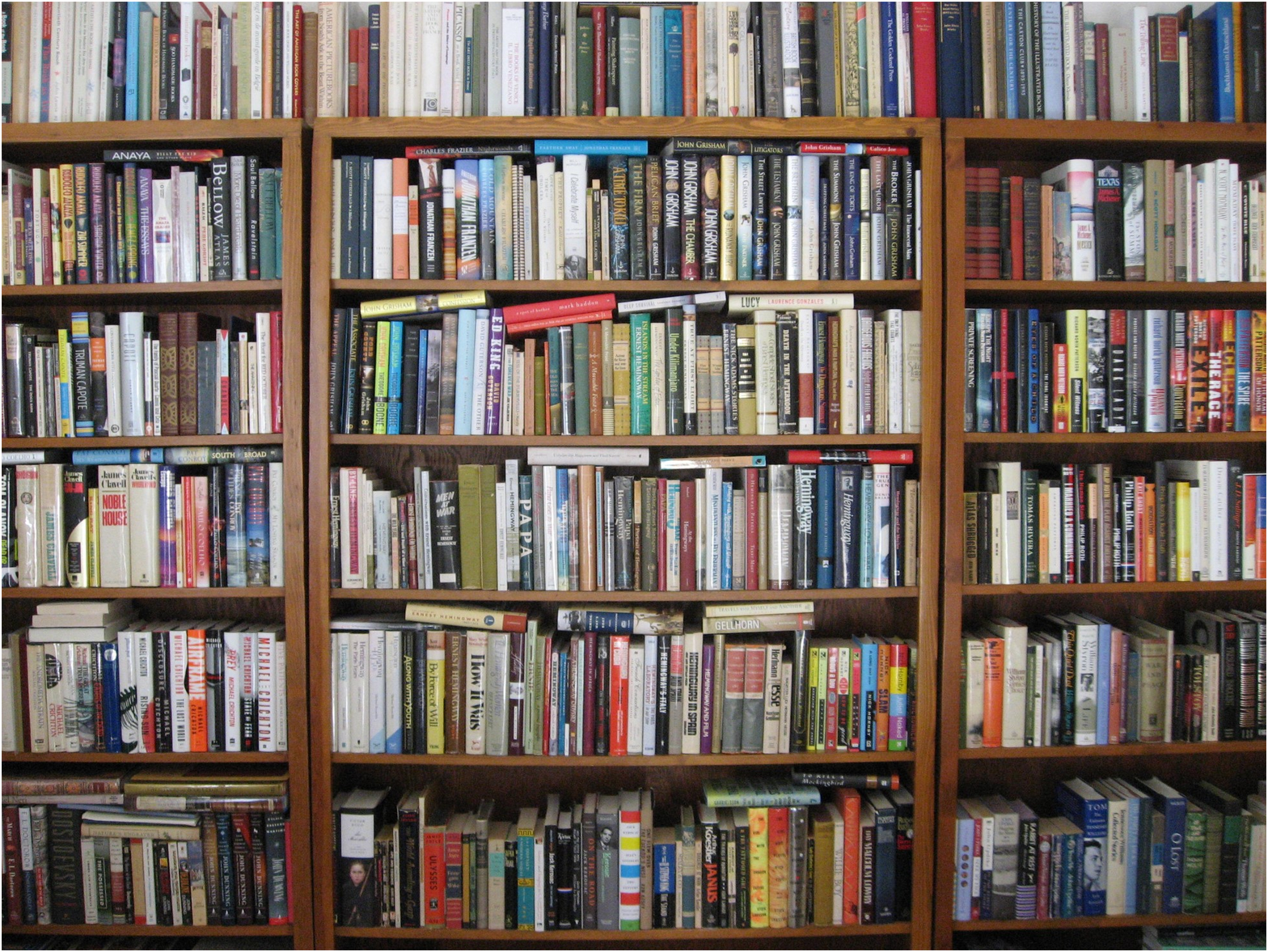}
		\label{fig:scn6}
	}
    \subfigure[Scene 4]{
		\includegraphics[width=0.22\columnwidth]{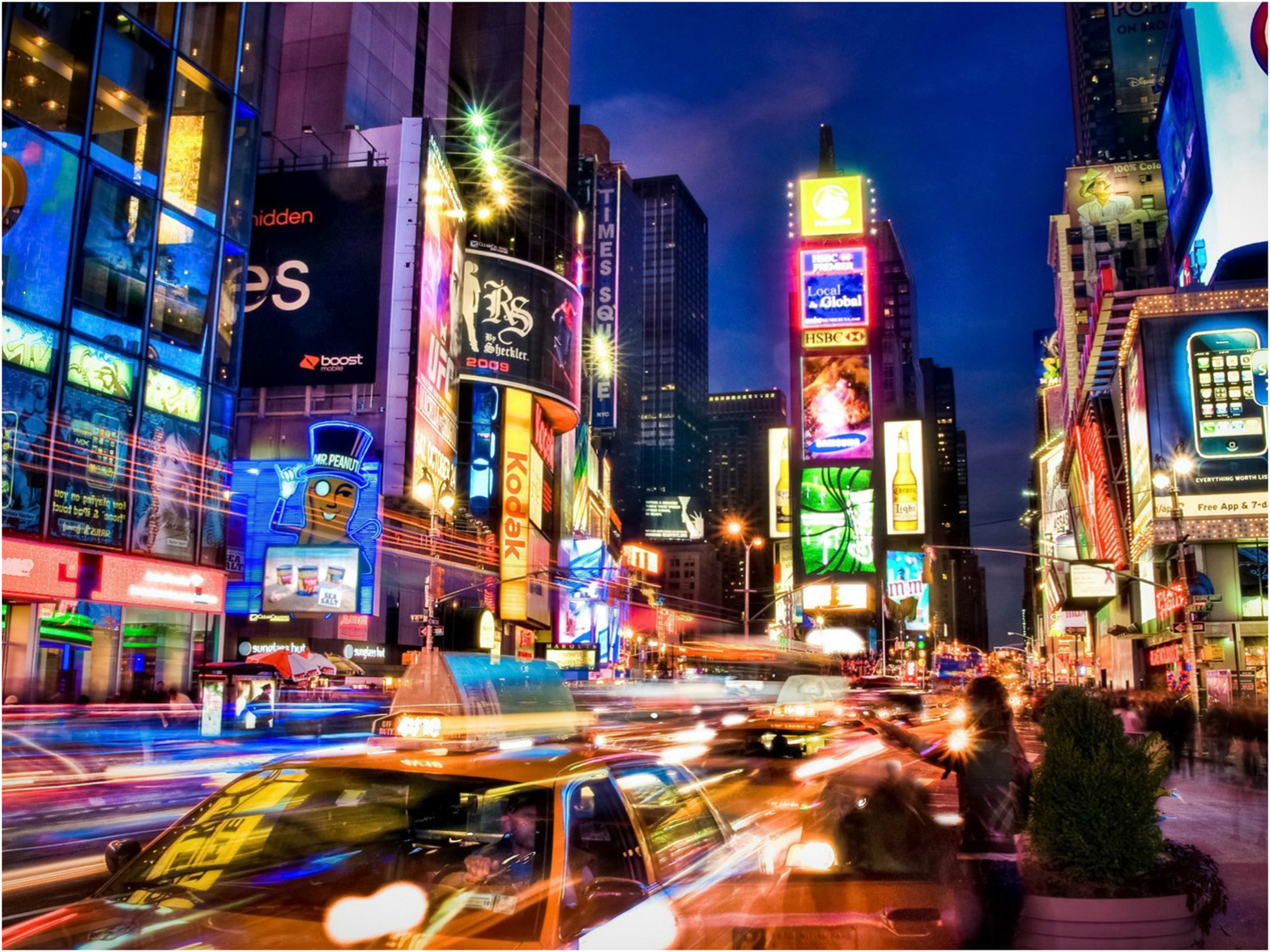}
		\label{fig:scn8}
	}
	\subfigure[Scene 5]{
		\includegraphics[width=0.22\columnwidth]{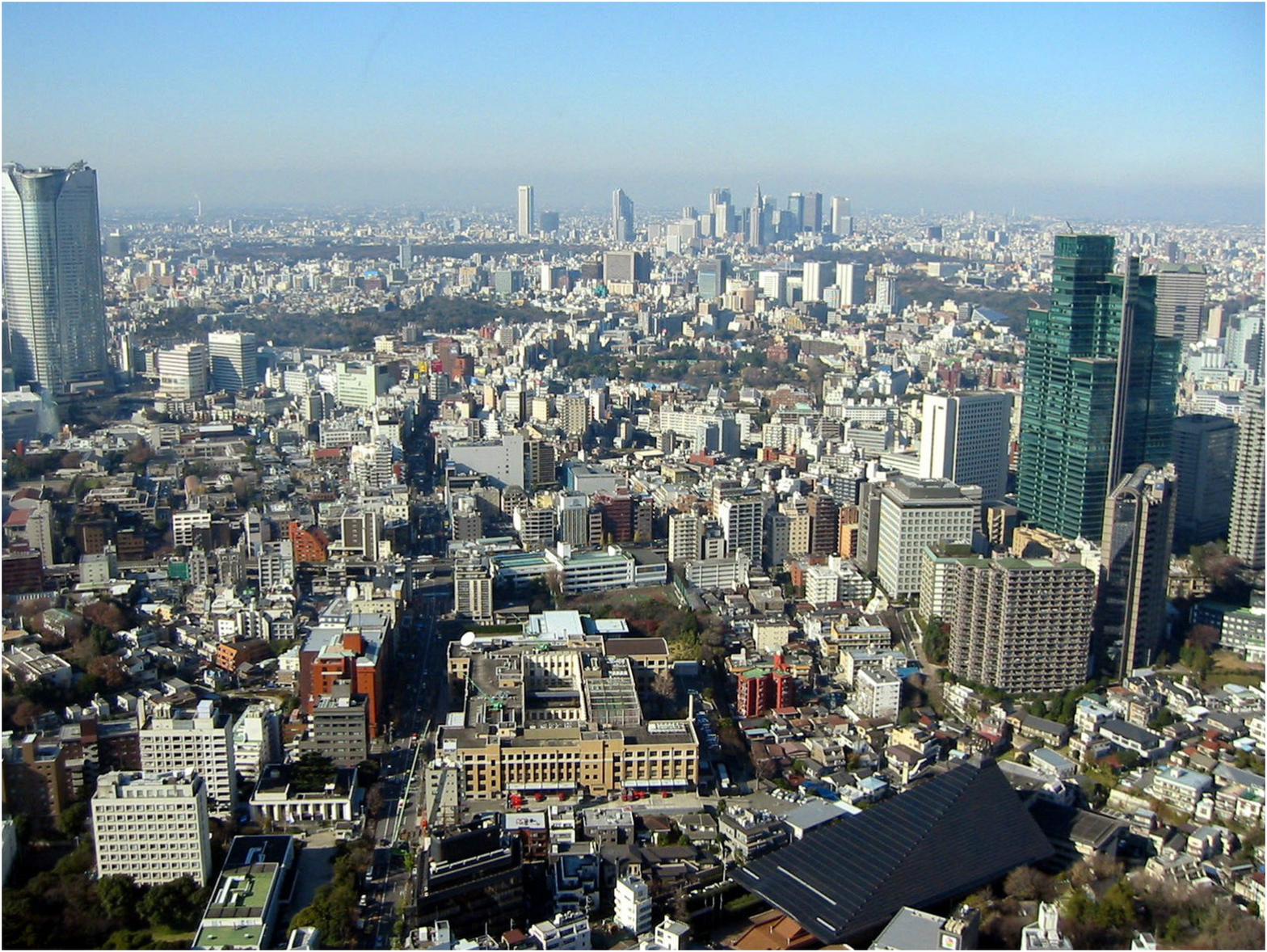}
		\label{fig:scn0}
	}
    \subfigure[Scene 6]{
		\includegraphics[width=0.22\columnwidth]{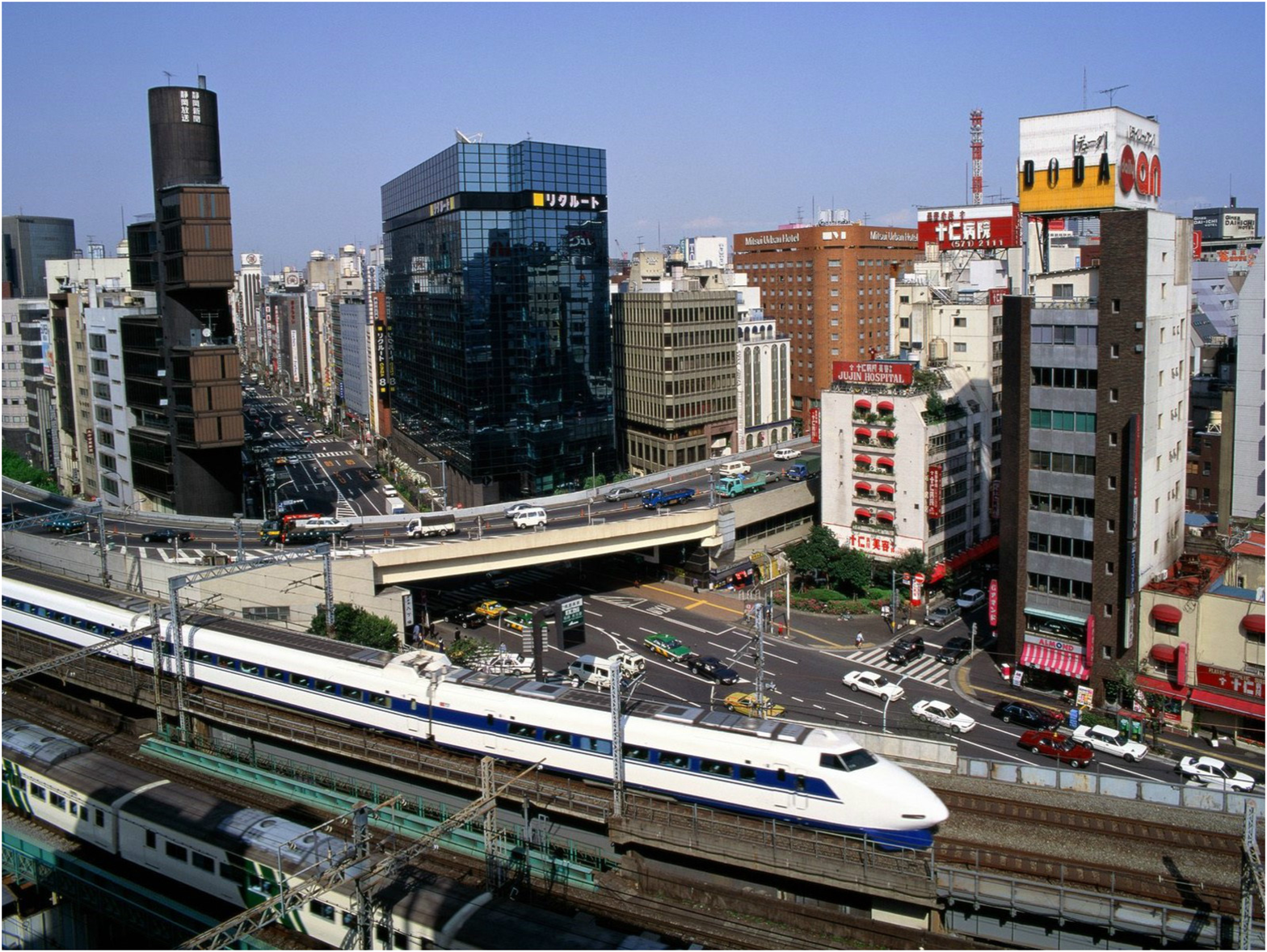}
		\label{fig:scn1}
	}
	\subfigure[Scene 7]{
		\includegraphics[width=0.22\columnwidth]{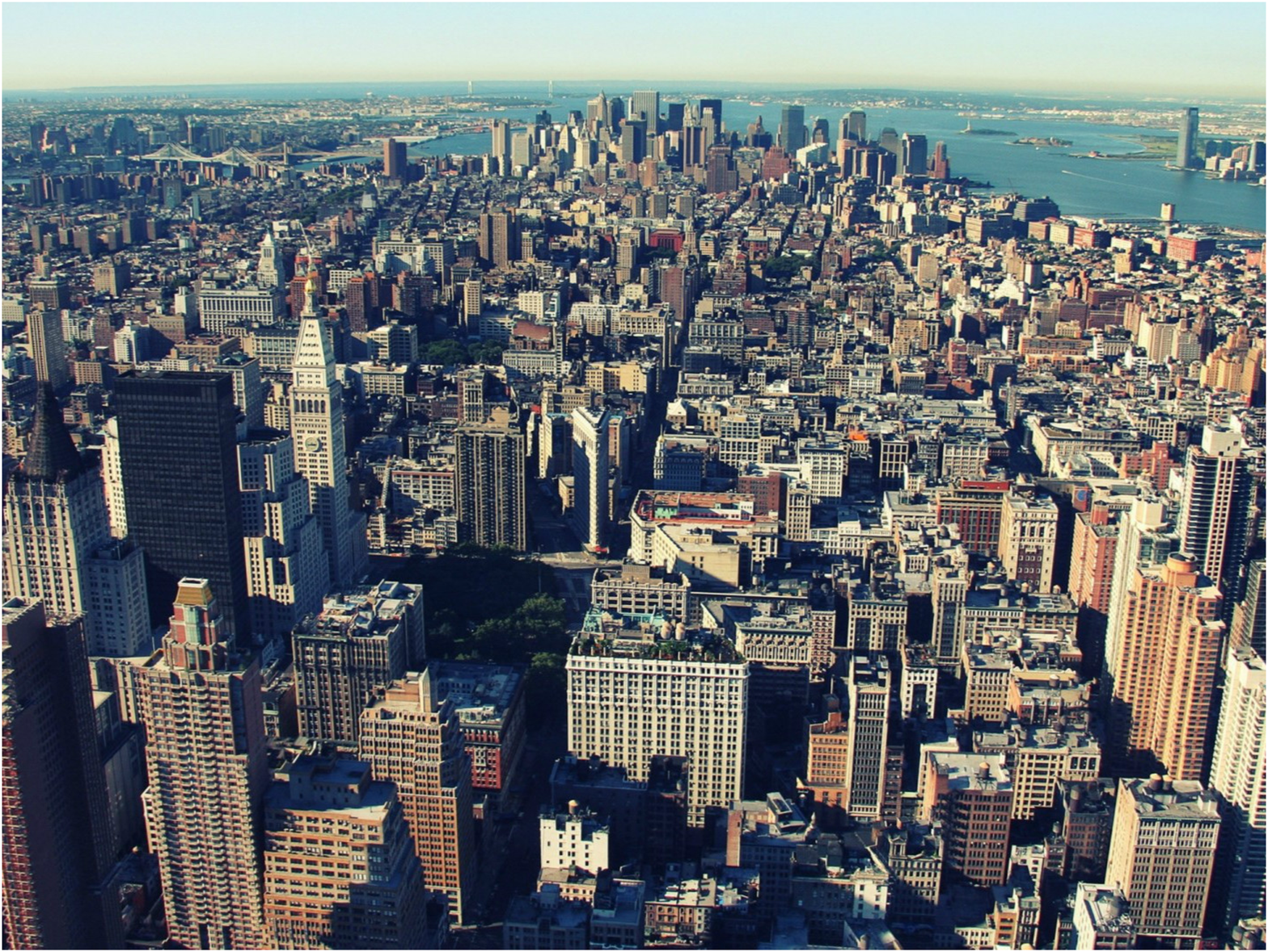}
		\label{fig:scn7}
	}
	\caption{Seven different backgrounds for the experiments. These include indoor and outdoor scenes in both day and night.}
	\label{fig:scn}
\end{center}
\end{figure}

In the experiments, we used a light-field camera with $5 \times 5$ viewpoints (ProFusion 25, Viewplus Inc.) to acquire the images. We placed the target objects about 50 cm from the camera, with the background a further 100 cm behind the objects.
We captured seven transparent objects (shown in Fig. \ref{fig:obj}) with seven different background scenes (shown in Fig. \ref{fig:scn}). The backgrounds include indoor scenes such as a library and outdoor scenes such as a city backdrop seen through a window.

As mentioned in Sec. \ref{sec:LFC_LFD}, we use an optical flow algorithm to obtain the corresponding points $p,p'$ and $p''$ in the central viewpoint $view(0,0)$ and the other viewpoints $view(s,t)$.
We utilize the optical flow algorithm proposed in \cite{LDOF_pami}, which integrates descriptor matching into variational motion estimation. Although this optical flow algorithm is very accurate, it cannot deal with textureless regions, and such areas will cause problems when the matching is not correct. For this reason, we remove those textureless regions for which the squared horizontal intensity gradient averaged over a square window of a given size is below a given threshold \cite{scharstein2002taxonomy}. The parameters used in all experiments are fixed to the same values. We determine the parameters $\alpha$, $\beta$, $\gamma$ based on the preliminary experiments, and set $\alpha=70$, $\beta=4.5$, $\gamma=4.5$ which are suitable for our dataset. Parameter $a$ decide the thresholding is hard or soft, $b$ is determined by the level of least-squares error, and $\tau$ is related to the accuracy of optical flow and image resolution. We set $a=0.5$, $b=5$ and $\tau=8$, which are suitable for our case.

We compare our segmentation results with those from LF-linearity thresholding and the finding glass method.
For the thresholding method, we simply filter out the Lambertian background by removing feature points whose least-squares error $E(u,v)$ is below a certain threshold, i.e., $E(u,v)<th$. In our experiments, we set $th$=5 which is same to $b$. For the finding glass method, we implemented the method described in \cite{findingGlass} and applied to the center view of our dataset.

\begin{figure*}[t]
	\centering
	\includegraphics[width=1.8\columnwidth]{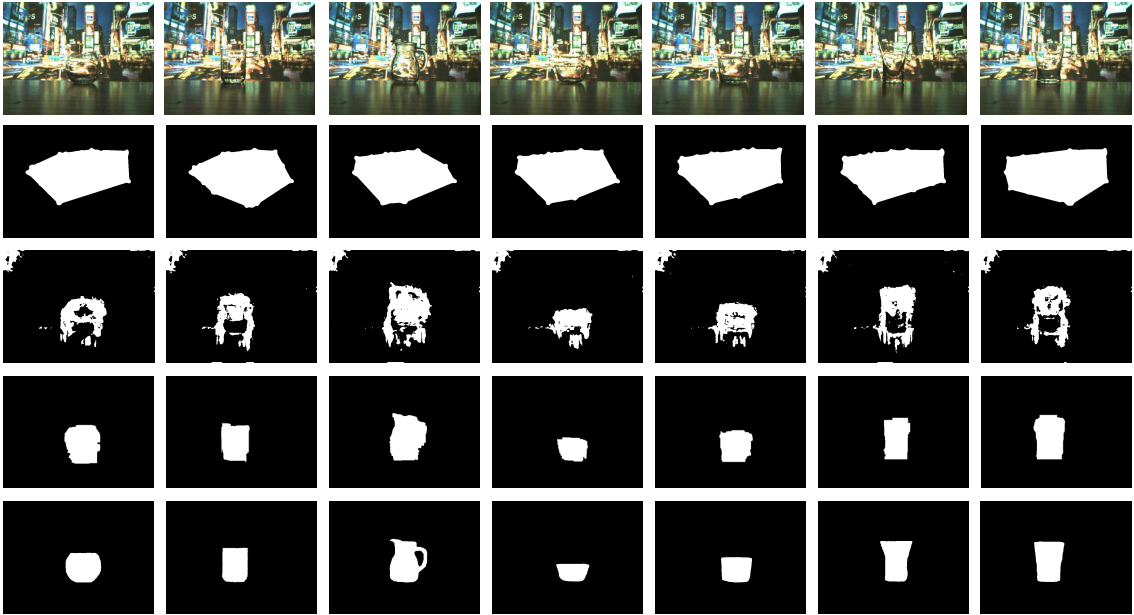}
	\caption{Comparison of segmentation results for the same scene with different objects. The 1st row shows the image from the central viewpoint. The 2nd, 3rd, and 4th rows show output from the finding glass, LF-linearlity thresholding, and proposed TransCut methods, respectively. The last row shows the manually labeled ground truth.}
    \label{fig:result_scn4}
\end{figure*}

\begin{figure*}[t]
	\centering
	\includegraphics[width=1.8\columnwidth]{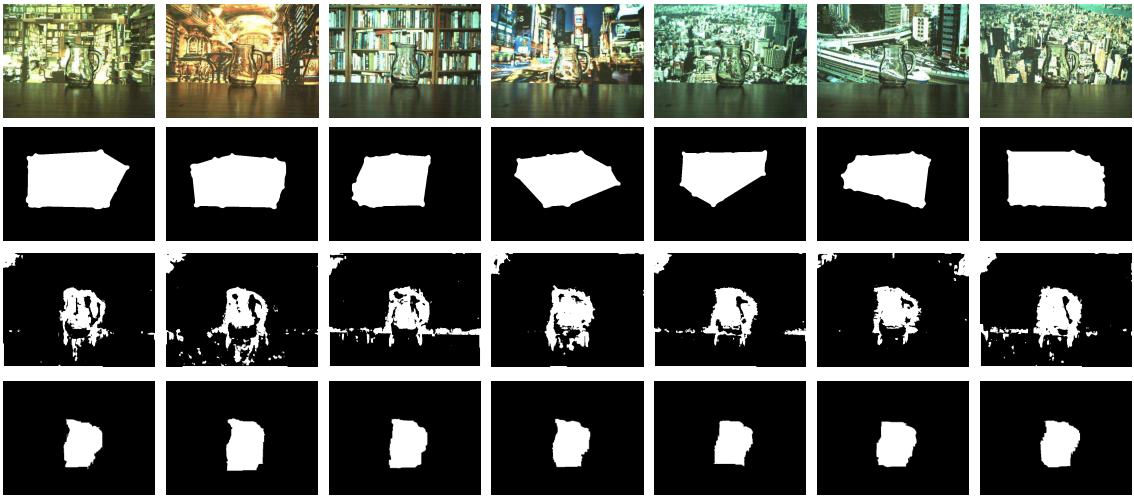}
	\caption{Comparison of segmentation results for the same object in different scenes. The 1st row shows the image from the central viewpoint. The 2nd, 3rd, and 4th rows show output from the finding glass, LF-linearity thresholding, and proposed TransCut methods, respectively.  We refer to the 3rd object in the last row of Fig. \ref{fig:result_scn4} for the ground truth.}
    \label{fig:result_obj5}
\end{figure*}

Figure \ref{fig:result_scn4} shows the results for the same scene with different objects, and Fig. \ref{fig:result_obj5} shows the segmentation results for the same object with different backgrounds.
We can see that, simply LF-linearity thresholding will result in holes inside the target object at points where the light field is nearly linear, and mismatched regions from outside will be included in the object.
The finding glass method falsely detected the rich texture background as glass, since this method is not suitable for rich texture images, which is mentioned as the limitation in the paper.
The proposed TransCut method gives very stable results for various objects in different scenes.

We determined the ground truth by manually labeling all pixels, and quantitatively compared the segmentation results. This comparison is tabulated in Table \ref{tab:quantitative_cmp}.
We have used the F-measure to compare the performance of each algorithm. This metric is the harmonic mean of the precision (Pr) and recall (Re), i.e.,
\begin{equation}
F=\frac{2*Pr*Re}{Pr+Re},
\end{equation}
where $Re= TP/(TP+FN)$, and $Pr= TP/(TP+FP)$\\
(TP=True Positive, FN=False Negative, FP=False Positive).

\begin{table}[t]
	\begin{center}
	\begin{tabular}{|c||c|c|c|}
		\hline
		& F-measure & Recall & Precision \\
        \hline
        Finding glass & 0.30 & 0.82 & 0.19 \\
		LF-linearity thresholding & 0.48 & 0.70 & 0.37 \\
        Proposed method & 0.85 & 0.96 & 0.77 \\
	\hline
	\end{tabular}
    \end{center}
    \caption{Quantitative comparison of three methods. The results are averaged over the single object dataset with 7 objects and 7 scenes.}
    \label{tab:quantitative_cmp}
\end{table}
We also compare the results calculated by different number of viewpoints. We reduced the viewpoints to the central view with 4 far corner views, and uniformly distributed $3 \times 3$ views with larger disparity. The results are shown in Table \ref{tab:viewpt_cmp}. We can see that the performance decreases when the viewpoints become fewer, because fewer viewpoints are more vulnerable to the noise.

\begin{figure}[t]
	\centering
	\includegraphics[width=.7\columnwidth]{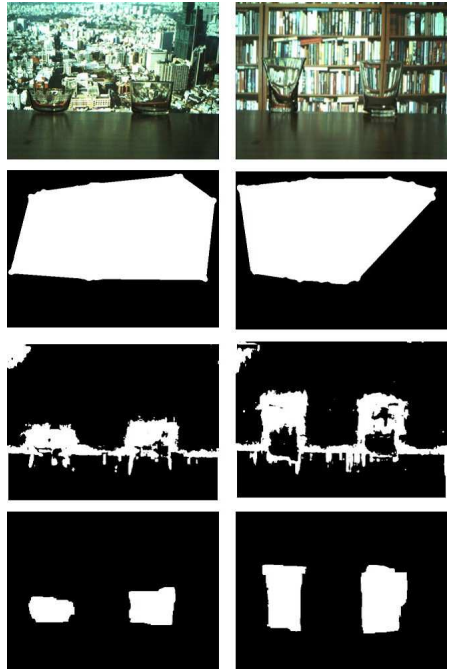}
	\caption{Comparison of segmentation results for multiple objects in different scenes. The 1st row shows the image from the central viewpoint. The 2nd, 3rd, and 4th rows show output from the finding glass, LF-linearity thresholding, and proposed TransCut methods, respectively.}
    \label{fig:double_objs}
\end{figure}

\begin{table}[t]
	\begin{center}
	\begin{tabular}{|c||c|c|c|}
        \hline
		& F-measure & Recall & Precision \\
        \hline
        5 Viewpoints & 0.76 & 0.75 & 0.78 \\
		9 Viewpoints & 0.82 & 0.85 & 0.79 \\
       25 Viewpoints & 0.85 & 0.96 & 0.77 \\
        \hline
	\end{tabular}
    \end{center}
    \caption{Comparison of different viewpoints number.}
    \label{tab:viewpt_cmp}
\end{table}
The results of experiments including multiple objects are shown in Fig. \ref{fig:double_objs}. These images show that the proposed method is effective when there is more than one object in the scene, whereas the other two methods do not produce good results in such scenarios. Further results can be found in our supplementary material.

\begin{figure}[t]
	\begin{center}
	\includegraphics[width=0.7\columnwidth]{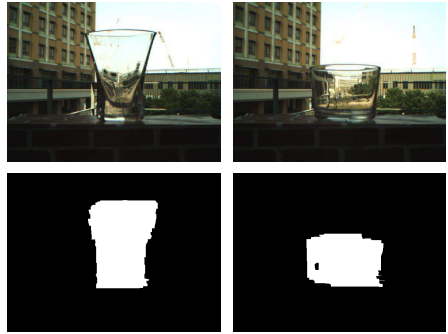}
    \caption{Results for real scene.}
    \label{fig:real_scene}
\end{center}
\end{figure}

Moreover, we also conduct some experiments with real scene. We can see that our method works though it is not perfect in Fig. \ref{fig:real_scene}.
\section{Conclusion}
In this paper, we have proposed TransCut which is the method for the segmentation of transparent objects. Unlike conventional methods, our technique does not rely on color information to distinguish the foreground and background. We have used LF-linearity and occlusion detector in 4D light field space for describing a transparent object, and designed an appropriate energy function utilizing the LF-linearity and occlusion for pixel labeling by graph-cut. The results show that our method produces stable results with various objects in different scenes.

There are several future directions we are planning to explore.
Our dataset was captured by camera array where the camera baseline is large and viewpoint number is few. An straightforward future step is to apply our algorithm to the light field image captured by Lytro camera.
The current results are not yet perfect, as our assumptions produce some limitations. We intend to overcome these limitations in future work, and extend to more flexible environment and other non-Lambertian objects.
\section*{Acknowledgement}
This research was supported by Grant-in-Aid for Scientific Research (A) No. 25240027.
{\small
\bibliographystyle{ieee}
\bibliography{transcut_short}
}
\end{document}